\documentclass[journal]{IEEEtran}
\usepackage{ulem}
\usepackage{xcolor}
\usepackage{color}
\usepackage{colortbl}
\usepackage[noadjust, nocompress]{cite}
\usepackage[pdftex]{graphicx}
\usepackage{amssymb}
\usepackage{amsfonts}
\usepackage{url}
\usepackage{multirow}
\usepackage{tablefootnote}
\usepackage{threeparttable}
\usepackage{booktabs}
\usepackage{pifont}
\definecolor{citecolor}{RGB}{66,168,235}
\definecolor{linkcolor}{RGB}{255,0,0}
\definecolor{urlcolor}{RGB}{0 0, 0}
\definecolor{greencolor}{RGB}{0.204,0.659,0.325}
\usepackage{hyperref}
\hypersetup{colorlinks=true,citecolor=citecolor,linkcolor=linkcolor,urlcolor=urlcolor}
\newcommand{\cmark}{\ding{51}}%
\newcommand{\xmark}{\ding{55}}%
\definecolor{cyan}{cmyk}{.3,0,0,0}
\newcommand{\MYhref}[3][magenta]{\href{#2}{\color{#1}{#3}}}%

\usepackage{tikz}
\usepackage{textcomp}
\usepackage{lipsum}

\newcommand\copyrighttext{%
  \footnotesize \textcopyright 2023 IEEE. Personal use of this material is permitted.  Permission from IEEE must be obtained for all other uses, in any current or future media, including reprinting/republishing this material for advertising or promotional purposes, creating new collective works, for resale or redistribution to servers or lists, or reuse of any copyrighted component of this work in other works.
  DOI: \href{<http://tex.stackexchange.com>}{10.1109/TGRS.2023.3268232}}
\newcommand\copyrightnotice{%
\begin{tikzpicture}[remember picture,overlay]
\node[anchor=south,yshift=6pt] at (current page.south) {\fbox{\parbox{\dimexpr\textwidth-\fboxsep-\fboxrule\relax}{\copyrighttext}}};
\end{tikzpicture}%
}

\begin{document}
\title{CMID: A Unified Self-Supervised Learning Framework for Remote Sensing Image Understanding}
%
%
%

\author{Dilxat~Muhtar,
        Xueliang~Zhang,~\IEEEmembership{Member,~IEEE,}
        Pengfeng~Xiao,~\IEEEmembership{Senior~Member,~IEEE},
        Zhenshi~Li,
        and~Feng Gu
\thanks{The authors are with the Jiangsu Provincial Key Laboratory of Geographic
Information Science and Technology, Key Laboratory for Land Satellite Remote Sensing Applications of Ministry of Natural Resources, School of Geography and Ocean Science, Nanjing University, Nanjing 210023, China (e-mail: 502022270062@smail.nju.edu.cn; zxl@nju.edu.cn; xiaopf@nju.edu.cn).}

\thanks{Corresponding Author: X. Zhang. This work was supported by the National Natural Science Foundation of China (Grant No. 42071297, 41871235).}
}

\maketitle
\copyrightnotice
\begin{abstract}
Self-supervised learning (SSL) has gained widespread attention in the remote sensing (RS) and earth observation (EO) communities owing to its ability to learn task-agnostic representations without human-annotated labels. Nevertheless, most existing RS SSL methods are limited to learning either global semantic separable or local spatial perceptible representations. We argue that this learning strategy is suboptimal in the realm of RS, since the required representations for different RS downstream tasks are often varied and complex. In this study, we proposed a unified SSL framework that is better suited for RS images representation learning. The proposed SSL framework, Contrastive Mask Image Distillation (CMID), is capable of learning representations with both global semantic separability and local spatial perceptibility by combining contrastive learning (CL) with masked image modeling (MIM) in a self-distillation way. Furthermore, our CMID learning framework is architecture-agnostic, which is  compatible with both convolutional neural networks (CNN) and vision transformers (ViT), allowing CMID to be easily adapted to a variety of deep learning (DL) applications for RS understanding. Comprehensive experiments have been carried out on four downstream tasks (i.e. scene classification, semantic segmentation, object-detection, and change detection) and the results show that models pre-trained using CMID achieve better performance than other state-of-the-art SSL methods on multiple downstream tasks. The code and pre-trained models will be made available at \MYhref{https://github.com/NJU-LHRS/official-CMID}{https://github.com/NJU-LHRS/official-CMID} to facilitate SSL research and speed up the development of RS images DL applications.
\end{abstract}

\begin{IEEEkeywords}
self-supervised learning, remote sensing pretraining, deep learning, contrastive learning, masked image modeling
\end{IEEEkeywords}

%
\IEEEpeerreviewmaketitle

\section{Introduction}

\IEEEPARstart{D}{eep} learning (DL) has gained recognition for its ability to automatically extract representations that accurately reflect the intrinsic properties of images, making it a successful tool in understanding remote sensing (RS) images~\cite{saleem2021automation,zheng2019cropdeep,cheng2016survey,li2020object}.
While supervised DL paradigms have witnessed great success in RS applications~\cite{8113128,MA2019166,yuan2020deep,ball2017comprehensive}, they require large-scale, high-quality annotated data, which is difficult to obtain as accurately annotating RS images is time-consuming and requires fine domain expertise.
Moreover, the reliance on labeled data can also limit the ability of the model to generalize to new or unseen data, which can be a challenge in RS applications where the environment is constantly changing~\cite{wang2022self}. In contrast, it is much easier to obtain large amounts of unlabeled RS images due to the growing number of satellites used for earth observation (EO). This limited labeling capability, along with the massive unlabeled images, makes representation learning on RS images highly relevant for self-supervised learning (SSL).

\begin{figure}[t]
    \centering
    \includegraphics[width=1\linewidth]{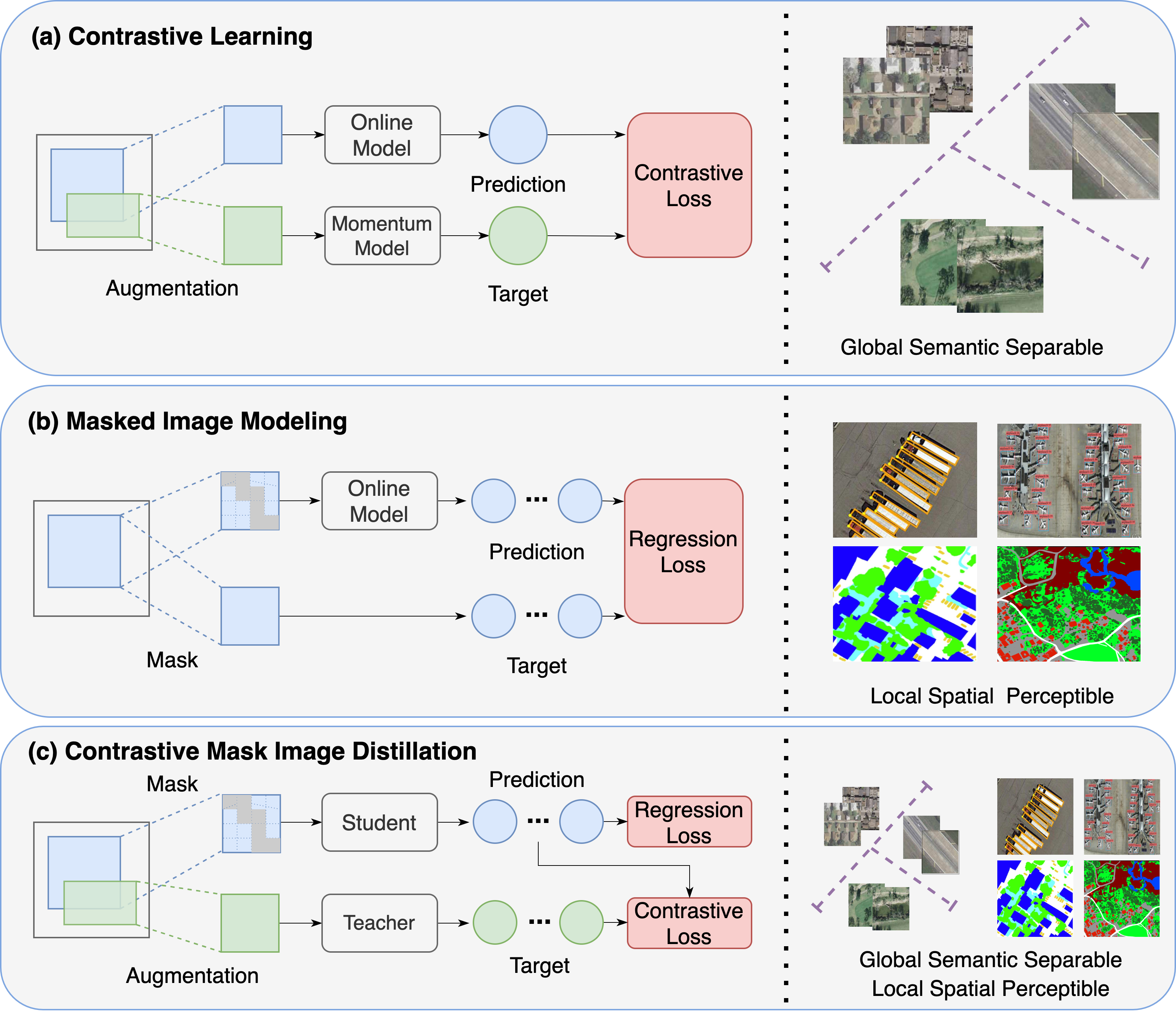}
    \caption{Comparisons between different self-supervised learning (SSL) paradigms. 
    (a)~Contrastive learning (CL) distinguishes global representations of positive pairs from those of negative pairs, making the pre-trained model with strong global semantic separability but less local spatial perceptibility.
    (b)~Masked image modeling (MIM) aims at learning local spatial relations for fulfilling the reconstruction task, resulting in the learned representations to be more local spatial perceptible but less global semantic separable.
    (c)~The proposed Contrastive Mask Image Distillation (CMID) aims to take the best of CL and MIM  and learns representations with both global semantic separability and local spatial perceptibility in a unified way.}
    \label{leadingfigure}
\end{figure}

SSL methods learn from vast amounts of unlabeled data by leveraging the structure present in the data itself to create supervised signals~\cite{jing2020self,chen2020improved,misra2020self}.
This possibility to train deep learning models without human-annotated labels, combined with the impressive results achieved by SSL methods on natural images~\cite{grill2020bootstrap,chen2020improved}, has prompted numerous investigations into adopting SSL for RS images~\cite{sun2022ringmo,manas2021seasonal,ayush2021geography}.
Contrastive learning (CL) and masked image modeling (MIM) are currently the most widely-adopted SSL methods in the domain of RS~\cite{jing2020self}.
CL involves bringing closer the representations of semantically similar inputs (such as two augmented views of the same image) and separating those of different images~\cite{jaiswal2020survey}.
The efficiency of CL in learning separable representations and the availability of naturally semantically similar samples within RS images, such as multi-temporal data or data acquired from multiple sensors at a single location, have established CL as the leading SSL paradigm in RS~\cite{kang2020deep,swope2021representation}.
The flexibility of creating contrastive signals in CL has also driven significant efforts to incorporate unique information about RS images into CL method, resulting in the acquisition of more specialized representations.
For example, multi-temporal RS images and geographic information have been combined to generate positive pairs, thus allowing the model to learn spatio-temporal invariant representations~\cite{manas2021seasonal, ayush2021geography}.
Global land cover products and multi-modal RS data have also been integrated with CL method to learn more generalized representations~\cite{li2021geographical,chen2021self}.
Most recently, the Masked Image Modeling (MIM) SSL method, which learns representations by reconstructing the missing information from the masked image has gained popularity in the field of RS~\cite{yuan2022sits,cong2022satmae,scheibenreif2022self,sun2022ringmo,wang2022advancing} with the success of vision transformer (ViT)~\cite{dosovitskiy2020image} in RS images~\cite{wang2022advancing,wang2022empirical,aleissaee2022transformers} and the remarkable breakthrough made by MIM in natural images~\cite{he2022masked,xie2022simmim}.
The recent study~\cite{wang2022advancing} applied MIM methods on a customized, large-scale plain ViT model for RS images and conducted extensive experiments on various RS downstream tasks to explore the effectiveness of the MIM learning approach. 
 The results validate the capability of the MIM method in learning rich representations and demonstrate the potential of MIM for representation learning in the field of RS.

Although the existing SSL methods have gained remarkable achievements in representation learning for RS images, they are restricted to utilizing a single SSL method, either CL or MIM.
As a result, the learned representations suffer from a limitation in incorporating both global semantic separability and local spatial perceptibility due to the inherent limitations of CL or MIM methods~(Fig.~\ref{leadingfigure}).
Specifically, CL utilizes whole-image representations to capture inter-image relationships by integrating global semantic information, which enhances the separation of representations between different categories or semantics in the feature space, leading to representations with global semantic separability and improved performance in tasks such as linear probing and classification~\cite{zhou2022mimco,huang2022contrastive}. 
Hence, as CL merely concentrates on the inter-image semantic relationships and ignores the intra-image structure, it fails to perceive the semantic information at different spatial locations within the images and thus perform poorly in dense prediction tasks such as object detection and semantic segmentation~\cite{tao2022siamese}.
On the other hand, MIM learns the intra-image structure and perceives the contextual information of each pixel within an image, resulting in the learned representations possess local spatial perceptibility with the expense of global semantic separability. 
As a result, MIM pre-trained models perform well in dense prediction tasks but are not as effective in the classification task.~\cite{xie2022simmim, he2022masked}.
We argue that the most appropriate SSL method for RS images should enhance the performance of the model in various downstream tasks, which means that the learned representations should be both global semantic separable and local spatial perceptible since the required representations for different RS downstream tasks are often varied. 
For example, RS images are characterized by multi-resolution, multi-season, and multi-spectral, making them vary significantly in time and location~\cite{cheng2017remote}. Therefore, a global semantic separable representation is necessary to classify different scenes. On the other hand, the complicated object distribution and spatial arrangement, along with the variations in both object scale and style under different capturing scenes of RS images~\cite{li2021gsdet}, also require a robust and local spatial perceptible representation in order to accurately detect the objects in these images. 

The motivation of learning representations with both global semantic separability and local spatial perceptibility inspires us to design a unified SSL framework that takes the advantages of both CL and MIM.
 However, it is challenging to fully exploit the advantages of both methods since CL and MIM often use different learning objectives, data augmentations, and model architectures. The different learning objectives of these two methods result in capturing different levels of abstraction and semantics~\cite{huang2022contrastive}, and simply aligning them in the same feature space will lead to semantic confusion and misalignment, ultimately degrading performance on downstream tasks. Moreover, RS images are often known for their multi-object characteristics~\cite{muhtar2022index}, and the operation of masking out a large portion of images in MIM may vanish the small objects~\cite{sun2022ringmo} and introduce noise that is not originally present in the images~\cite{liu2022mixmim}, resulting in incompleteness in semantic meaning for the images. Consequently, performing CL on these semantic incomplete images will hinder the learning of separable representations. 

In this study, we propose Contrastive Mask Image Distillation (CMID) to overcome these challenges and learn both global semantic separable and local spatial perceptible representations in a unified way. CMID leverages teacher-student self-distillation architecture~\cite{zhang2019your,caron2021emerging} and incorporates the concept of multiple embedding sub-spaces~\cite{manas2021seasonal,xiao2020should} to take the best of CL and MIM. Specifically, the student encoder maps the masked images into latent embedding, while the teacher encoder encodes the full augmented images to ensure semantic integrity and provide contrastive supervision to guide the student. The latent embedding is then projected into different feature spaces to perform either reconstruction or discrimination tasks, which helps to prevent confusion between different levels of semantics. In the reconstruction task, the student learns representations that capture local spatial details by recovering the masked parts in both spatial and frequency domains. In the discrimination task, the student learns global semantic separable representations by discriminating the embedding of the student and teacher from those of other images. Additionally, we align the local semantics of the feature vectors at corresponding positions in the teacher's and student's inputs to overcome the semantic incompleteness caused by the large mask ratio and enhance the local separability of the learned representations. Finally, although the existing MIM methods are largely confined to ViT architectures and have been shown to suffer from information loss when applied to convolutional neural networks (CNN)~\cite{jing2022masked,liu2022mixmim}, CMID has been proven to be effective for both CNN and ViT architectures. This versatility allows for easy adaptation of CMID to a wide range of DL applications for RS images.

The main contributions of this study are as follows:
\begin{itemize}
    \item [(1)] We propose a unified SSL framework, CMID, that learns both global semantic separable and local spatial perceptible representations by combining the CL and MIM, which makes it suitable for a variety of RS downstream tasks. To our best knowledge, this is the first SSL framework that combines CL with MIM in the realm of remote sensing.
    \item [(2)] The proposed CMID method adopts a teacher-student self-distillation architecture to fully leverage the benefits of CL and MIM. This approach preserves semantic integrity and learns representations through distillation of both local and global semantics. Additionally, CMID incorporates the concept of multiple sub-spaces for latent embeddings to separate the different levels of semantics from CL and MIM, thus preventing semantic confusion.
    \item [(3)] Comprehensive experiments have been carried out on four RS downstream tasks (i.e. scene classification, semantic segmentation, object detection, and change detection) and the results show that models pre-trained using CMID achieve better performance than other state-of-the-art SSL methods on multiple downstream tasks. Moreover, these results demonstrate that CMID is effective for both CNN and ViT architectures.
\end{itemize}

\section{Related Work}
SSL is a representation learning paradigm that involves training a model to learn from unlabeled data by leveraging pre-text tasks to generate labels or supervision from the data itself~\cite{jing2020self,chen2020improved,misra2020self}. The goal of SSL is to learn general representations that can be used for various downstream tasks, such as classification~\cite{Stojnic_2021_CVPR} and semantic segmentation~\cite{muhtar2022index}. This section briefly discusses the two dominant learning frameworks of SSL~\cite{liu2021self}: contrastive learning (CL) and masked image modeling (MIM), and their applications to RS images.

\subsection{Contrastive Learning}
The core idea of CL is to pull together the representations of different views of the same image (positive samples) and push away that of the other images (negative samples)~\cite{oord2018representation,jaiswal2020survey}. 
SimCLR~\cite{chen2020simple}, which utilized a simple siamese network and a plentiful of data augmentation methods, was the first CL method to demonstrate competitive results compared to supervised learning. 
However, SimCLR directly used negative samples coexisting in the current batch and required a large batch size to ensure the diversity of the negative samples.
Considering this limitation, MoCo~\cite{chen2020improved} proposed using a large queue to cache negative samples such that it can take in more negative samples for CL and release the requirement of a large batch size of SimCLR. 
On the other hand, some CL methods have attempted to eliminate the reliance on negative samples through asymmetric design~\cite{grill2020bootstrap,chen2021exploring} or reducing redundancy~\cite{zbontar2021barlow}.
Specifically, BYOL~\cite{grill2020bootstrap} learned meaningful representations without negative samples by using an asymmetric architecture, where the online encoder predicts the output of a momentum encoder.
SimSiam~\cite{chen2021exploring} simplified the BYOL architecture by applying stop-gradient operations to replace momentum updates.
Barlow-Twins~\cite{zbontar2021barlow} learned representations by reducing the correlations of different images at a batch level.
On the other direction, SwAV~\cite{caron2020unsupervised} incorporated clustering into the CL framework to learn more discriminative representations by enforcing consistency between the assignments of the multi-augmented views of the same image. 
Finally, MoCo-v3~\cite{chen2021empirical} and DINO~\cite{caron2021emerging} adapted MoCo~\cite{chen2020improved} and BYOL~\cite{grill2020bootstrap} for pre-training the ViT~\cite{dosovitskiy2020image} network using CL methods.

CL pre-trained models have achieved great success in classification tasks due to their ability to learn global semantic separability during the pre-training stage. However, their transfer performance for dense prediction tasks, which require more fine-grained and local information, has been limited~\cite{wang2021dense}. In an effort to improve this performance, several recent studies have attempted to learn at a local scale by imposing consistency among pixels~\cite{xie2021propagate,o2020unsupervised}, feature maps~\cite{wang2021dense,liu2020self} or image regions~\cite{xu2021regioncl,xiao2021region}. Despite these efforts, this approach has been found to be prone to overfitting on the pre-training dataset and has limited performance on downstream tasks~\cite{van2021revisiting}.

\begin{figure*}[t!]
    \centering
    \includegraphics[width=0.9\linewidth]{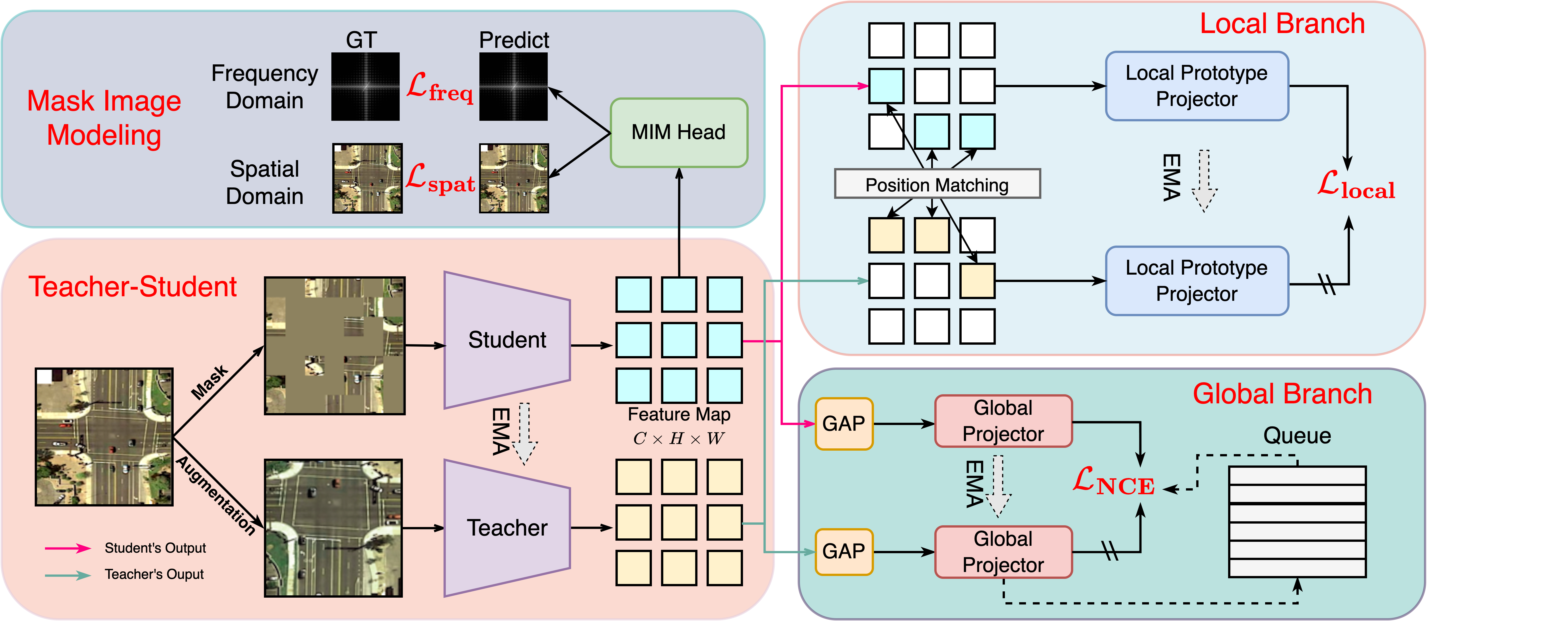}
    \caption{Overview of the proposed SSL framework, CMID, which adopts a teacher-student self-distillation architecture and consists of three branches.}
    \label{WholeModel}
\end{figure*}

\subsection{Masked Image Modeling}
Masked image modeling (MIM) learns rich visual representations via masked parts reconstruction by conditioning on visible context. 
Following BERT in natural language processing (NLP), iGPT~\cite{chen2020generative} compressed the image to a few pixels, and then directly predicted the masked pixels' color. However, the pixel-based method has a high pre-training computation cost. 
To reduce computation costs, BEiT~\cite{bao2021beit} proposed using a pre-trained discrete variational autoencoder~\cite{ramesh2021zero} as the visual tokenizer to encode masked patches and learn representations by reconstructing the masked visual tokens. 
Following this direction, PeCo~\cite{dong2021peco} proposed to inject perceptual similarity during the training of visual tokenizer to benefit MIM pre-trained representations.
Instead of using the pre-trained visual tokenizer, iBOT~\cite{zhou2021ibot} jointly trained the visual tokenizer and the target network via self-distillation.
MAE~\cite{he2022masked} improved the pre-training efficiency of MIM method by using an asymmetric encoder-decoder architecture to reconstruct raw images. 
Compared to MAE, MaskFeat~\cite{wei2022masked} and SimMIM~\cite{xie2022simmim} used simple linear layers as the decoder head instead of another transformer as in MAE. 
MaskFeat further enriched the learned representations by adopting histogram of gradient (HOG) as the reconstruction target instead of RGB values. 

MIM pre-trained models excel in transfer learning (TL) with fine-tuning~\cite{wei2022contrastive} and have demonstrated promising performance in dense prediction tasks~\cite{li2021benchmarking}. However, they tend to focus on learning the relationships within each image independently, without considering the inter-image relationships, leading the learned representations to be local semantic perceptible but less separable. In addition, most MIM methods have only been applied to ViT, while there have been some attempts~\cite{jing2022masked} to apply MIM to CNN, they have not performed as well as their ViT counterparts.
The underlying issue behind MIM with CNN lies in the fact that CNN performs sliding window on dense and regular inputs, hindering its ability to concentrate solely on the irregular visible regions to restore lost information.
Moreover, as the masked patches shifts the distribution of the original image and the convolution operation operates on both the visible and invisible patches, leading to further corruption of the image information~\cite{jing2022masked,tian2023designing}.

\subsection{Self-Supervised Learning in Remote Sensing}
The success of SSL in natural images representation learning has inspired numerous efforts to adapt SSL for RS images. 
CL is once the most popular SSL method in remote sensing community.
CMC~\cite{tian2020contrastive} was first adapted to RS images classification task and demonstrated the applicability of CL in RS images~\cite{Stojnic_2021_CVPR}. 
SimCLR~\cite{chen2020simple} was then incorporated with the idea of neighboring positive pairs to increase the diversity of the training data and learn smoothed representations for RS images~\cite{jung2021contrastive}.
SauMoCo~\cite{kang2020deep} applied MoCo~\cite{chen2020improved} CL method and defined a spatial augmentations criteria to uncover semantic relationships among land cover tiles. As a result, SauMoCo was able to characterize different unlabeled RS scenes.
Much work has also been done to incorporate additional information into CL to encourage models to learn more specific representations from RS images. 
Geography-aware MoCo~\cite{ayush2021geography} was able to close the gap between self-supervised and supervised learning on various RS images downstream tasks by leveraging spatially aligned images over time to construct temporal positive pairs in CL and using the geo-location regression task as an auxiliary task.
Similarly, SeCo~\cite{manas2021seasonal} used images from the same location at different temporal phases as positive pairs and applied the concept of multiple embedding sub-space to learn representations that encode the different variances and invariances. 
GeoKR~\cite{li2021geographical} utilized the global land cover products and regarded geographical location associated with each RS image as geographical knowledge to provide supervision for SSL pre-training.
Moreover, there are some works that focus on the application of CL methods to specific downstream remote sensing tasks. For example,  CL was integrated with change detection using multi-modal bitemporal scenes in an encoder-decoder architecture and achieved promising success in multisensor change detection~\cite{saha2021self}.
GLCNet~\cite{li2022global} and IndexNet~\cite{muhtar2022index} combined global and local contrast to learn more fine-grained representations for semantic segmentation of RS images.

Recently, the MIM method has been increasingly gaining popularity in the RS community owing to the success of the ViT in various RS downstream tasks. SITS-Former~\cite{yuan2022sits} proposed to learn spatio-spectral-temporal representations for time series classification by asking the model to regress central pixels of masked patches given an incomplete time series with some randomly masked patches. SatMAE~\cite{cong2022satmae} leveraged temporal and multi-spectral information in RS images to improve self-supervised pre-training with MIM. 
RingMo~\cite{sun2022ringmo} applied the MAE~\cite{he2022masked} method and designed a new mask strategy for self-supervised representation learning on a 3 million unlabeled RS images dataset. The fine-tuning results on various downstream tasks showed that the new mask strategy was more appropriate for RS images and the learned representations by RingMo were generalized well to various RS downstream tasks. 
MAE~\cite{he2022masked} was also applied to pre-train large vision models customized for RS, and the results on various RS downstream tasks demonstrated the effectiveness of MIM pre-training~\cite{wang2022advancing}. 

Although these approaches have achieved certain success, they are limited to learning either global semantic separable or local spatial perceptible representations. Moreover, these methods have prerequisites for the underlying architecture (i.e., CL methods rely on CNN, while MIM methods are confined to ViT). In contrast, the proposed method, CMID, is not only architecture-agnostic, but also learns representations with both global semantic separability and local spatial perceptibility, ensuring that the learned representations are generalizable enough to meet the requirements of various RS downstream tasks.

\section{Method}
\subsection{Overview}
The overall structure of the proposed CMID model is illustrated in Fig.~\ref{WholeModel}. 
CMID employs a teacher-student self-distillation architecture and consists of three branches: the MIM branch, the global branch, and the local branch. 
The MIM branch leverages the MIM method to learn local spatial perceptible representation. 
The global branch, using the CL method, focuses on learning global semantic separable representation. Meanwhile, the local branch aims to recover object-level information lost in the MIM branch through self-distillation. The entire network is designed to interact and balance between the learning signals from each branch through its teacher-student architecture. 
Specifically, given an image $x$, CMID first generates two views of $x$, referred to as masked image and augmented image, using either mask augmentation or random data augmentations, and feeds them into the student and teacher respectively. 
The teacher and the student share the same architecture (CNN or ViT). The parameters of the student are exponentially moving averaged (EMA) to the parameters of teacher. 
The student maps the masked image to latent embedding, while the teacher encodes the augmented image to preserve semantic integrity and provide contrastive supervision to guide the student. 
The MIM branch applies a MIM head to reconstruct the masked image using the latent embedding output by the student in both spatial and frequency domains.
The other two branches project the latent embedding to different feature spaces to avoid semantics confusion and leverage CL to learn separable representations by aligning the embedding of the student and teacher at both global and local scales. 
Specifically, the global branch utilizes the standard MoCo~\cite{chen2020improved} style CL method to discriminate global representations, while the local branch aligns the local semantics by enforcing assignment consistency between position-matched pairs over a set of prototypes to learn object-separable representations and avoid semantic incompleteness and misalignment. 

After pre-training using CMID with a large amount of unlabeled data, only the student is used for downstream tasks to evaluate the model performance. In the following subsections, we will provide a more detailed explanation of the different components of CMID.
 
 \subsection{Masked Image Modeling Branch}\label{maskimagemodeling}
 In the masked image modeling branch, we follow the practice of SimMIM~\cite{xie2022simmim} and adapt it to the representations learning of RS images with new careful designs.
 \subsubsection{SimMIM Framework}
 For the input image $x$, SimMIM first patchifies $x$ into non-overlapping image patches $x^p$ and randomly generates a mask $M$ to hide a portion of image patch. The masked patches are then replaced by a learnable mask token $\mathrm{[MASK]}$ which is initialized with a value of zero to indicate the presence of a missing patches to be predicted. The resulting image is referred to as $x_{mask}$:
 
 \begin{equation}
     x_{mask} = x^p \odot (1 - M) + \mathrm{[MASK]} \odot M,
 \end{equation}
 where $\odot$ is element-wise multiplication. The SimMIM encoder takes in the masked image $x_{mask}$ and generates a latent embedding. The lightweight one-layer MLP decoder (MIM head in Fig.~\ref{WholeModel}) then uses this embedding to output the predicted reconstruction image $x'$. Finally, an $\ell_1$ loss $\mathcal{L}_{spat}$ is employed on the masked pixels to update the model parameters:
 
 \begin{equation}
     \mathcal{L}_{spat} = \frac{1}{\Omega(x_m)} \parallel x_m - x'_m \parallel_1,
 \end{equation}
 where $x_m$ and $x_m'$ denote the sets of original and reconstructed values of the masked pixels, respectively; and $\Omega(\cdot)$ represents the number of elements in a set.
 \subsubsection{Adaptation for RS Images}
 Although the simplicity and effectiveness of SimMIM, there are some limitations must be taken into account when applying SimMIM into RS images. 
 One issue is that SimMIM replaces masked patches with the [MASK] token, but RS images are known for their multi-object characteristics~\cite{muhtar2022index} and the objects are usually densely distributed~\cite{sun2022ringmo}. The masking operation may cause the dense and small objects in the image to be lost~\cite{sun2022ringmo}, leading to incomplete semantic meaning and making image reconstruction more difficult~\cite{sun2022ringmo}. 
 Moreover, the zero-initialized mask token [MASK] is not originally presented in the image. Using it directly to replace the masked patches will cause inconsistency between the masked and augmented images, which will degrade the discriminativeness of the learned representations in the other two branches.

 To address the above issues, we adopt the strategy of using the mean spectral value to fill the masked patches, as suggested in \cite{li2022architecture}, and add the learnable mask token [MASK] to the patch embedding (the output of the patch embedding layer in ViT or the output of the stem layer in CNN), rather than simply replacing the masked patches. The $x_{mask}$ is now expressed as:
 \begin{equation}
     x_{mask} = x^P_{m} + \mathrm{MASK} \odot M,
 \end{equation}
 where $x^P_{m}$ denotes the set of image patches after the masked patches have been filled with mean spectral value. 
The consideration behind this design is that the spectral mean value is the direct center (DC) component of the image, resulting in minimum local statistics variation compared to the mask token replacement~\cite{li2022architecture}, and it also contains middle-level information about the image.
As a result, this masking strategy not only minimizes semantic discrepancy between the masked and augmented images but also reduces the difficulty of recovering the missing information by providing a middle-level information hint. 
 Moreover, we further mitigate the semantic incompleteness caused by the loss of small objects by aligning the local semantics between the student and teacher outputs, which will be described in Sec.~\ref{localbranch}. Finally, we also incorporate the focal frequency loss (FFL) $\mathcal{L}_{freq}$~\cite{jiang2021focal} in frequency domain to enforce consistency between the original and the reconstructed image, as this has been shown to be effective to learn representations with high-level semantics~\cite{he2022masked,li2022architecture}:
 \begin{figure}[t]
    \centering
    \includegraphics[width=1\linewidth]{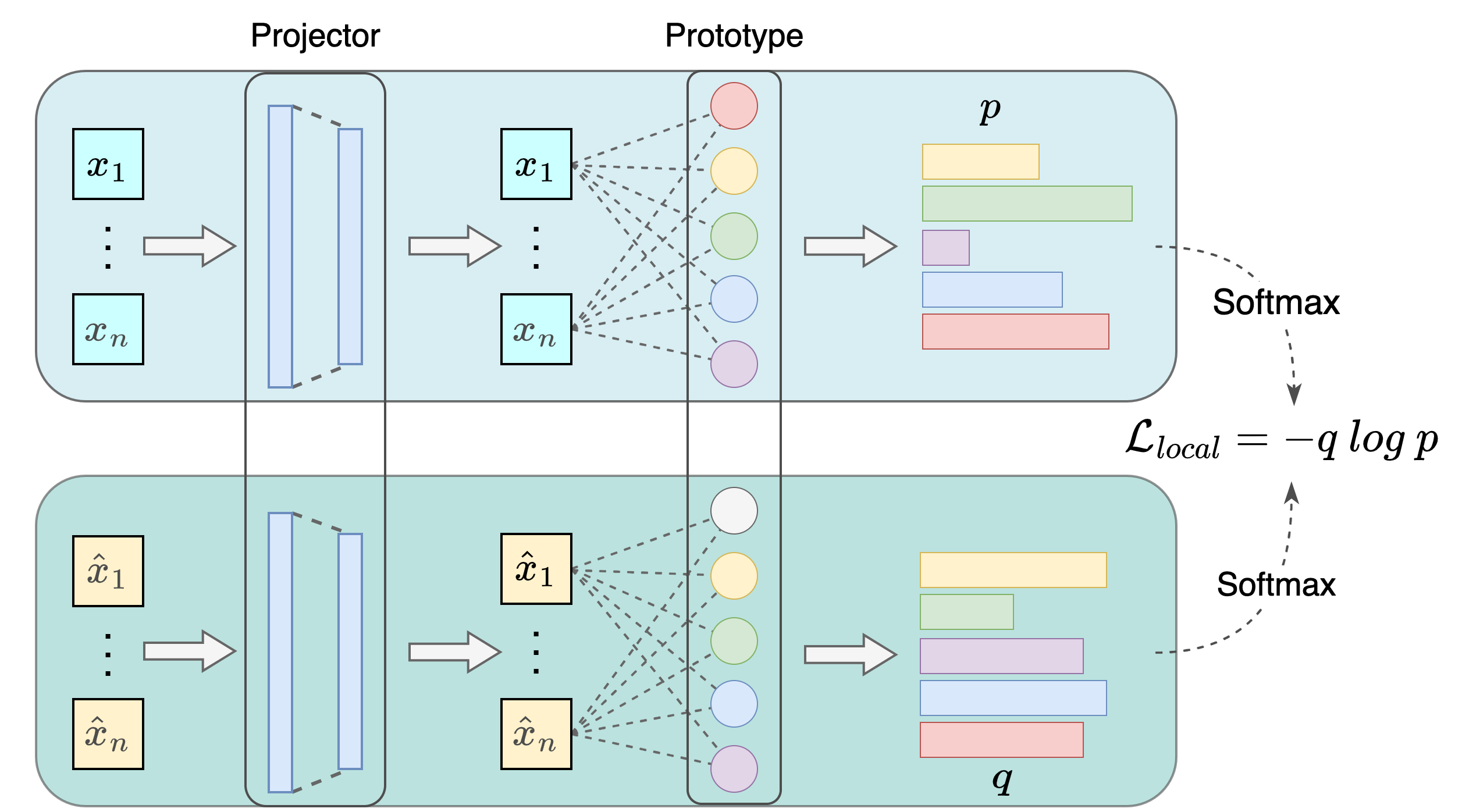}
    \caption{Detailed process of the local branch. The matched pairs are projected into another feature space and mapped to a set of learnable prototypes to obtain their respective similarities
    to these prototypes. The cross entropy loss is then used to minimize the difference in the distributions of these similarities.}
    \label{localbranchfigure}
\end{figure}
\begin{equation}
    \mathcal{L}_{freq} = \frac{1}{N} \sum_{c=1}^N FFL(x_c, x_c')
\end{equation}
where $N$ refers to the number of channels in the input image $x$; and the subscript $c$ indicates the specific channel of the image.
The detailed formulation of FFL please refer to~\cite{jiang2021focal,li2022architecture}.

\begin{figure*}[t]
    \centering
    \includegraphics[width=0.9\linewidth]{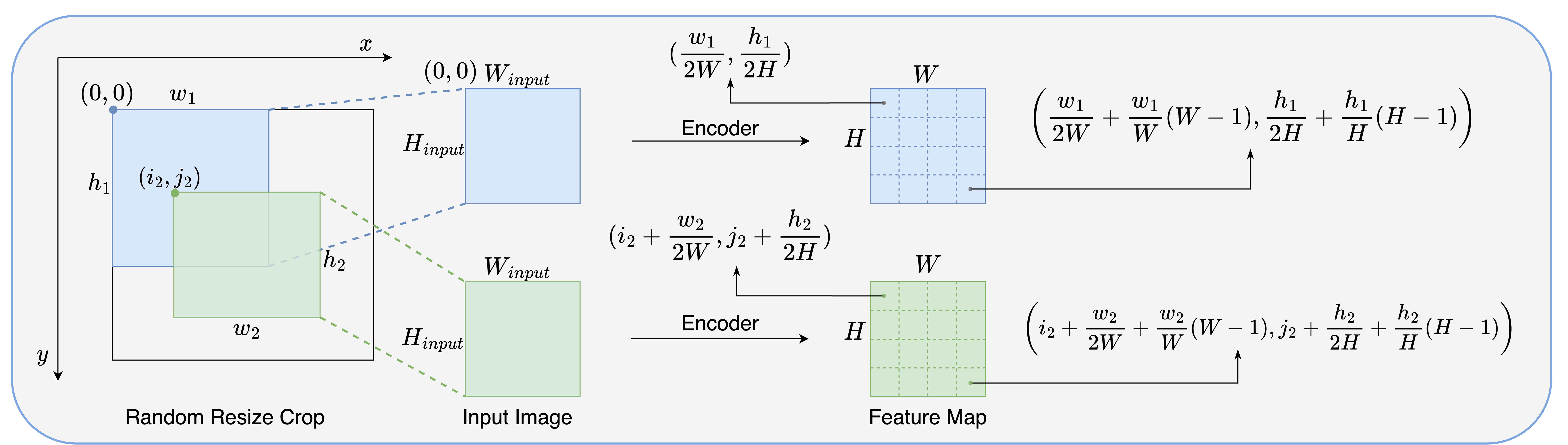}
    \caption{The detailed process of calculating the absolute position of each feature vectors. These absolute positions are calculated using the positional parameters of random resize crop augmentation and the input and output image sizes of the model.}
    \label{position}
\end{figure*}

 \subsection{Global Branch}
 In the MIM branch, the student learns local spatial perceptible representations by capturing fine-grained visual context and semantics, but loses
 global separability.
To bridge this gap, the student is trained to recover the global semantic content of the masked image and learn global semantic separable representations by aligning its representation with that of the teacher using a visual dictionary queue in the global branch. 
The outputs of the student and teacher serve as the query and key in the adopted MoCo~\cite{chen2020improved} CL method. 
Both the query and key are pooled using global average pooling (GAP) to obtain their respective global representations, which are then projected into another feature space to prevent confusion between different levels of semantics in the MIM and global branches.
Finally, we apply infoNCE loss~\cite{oord2018representation} to align their representations by discriminating them from other images:

\begin{equation}
    \mathcal{L}_{NCE} = - log \frac{exp(\langle q,k^{+}\rangle /\tau)}{\sum_{i=1}^Kexp(\langle q,k_{i}\rangle /\tau)}\label{infonce},
\end{equation}
where $\langle\cdot,\cdot\rangle$ indicates the cosine similarity; $q, k^{+}$ denote the projected global representations of the student and teacher; $\tau$ is a temperature scaling the distribution of distance; and $k_i$ is the member of the visual dictionary queue. The queue is built by the teacher's projected global representations and updated in each iteration~\cite{chen2020improved}.

 \subsection{Local Branch}\label{localbranch}
In Sec.~\ref{maskimagemodeling}, we discussed how the mask operation can cause semantic incompleteness by dropping dense and small objects in RS images. 
This incompleteness not only deteriorates the extraction of local spatial semantics in the MIM branch~\cite{sun2022ringmo}, but also influences the global branch due to the ambiguity it brings to the semantic meaning of the image~\cite{huang2022contrastive}. 
To address this issue, we further mitigate semantic incompleteness in the local branch by aligning the local semantics of the student and teacher.
This alignment is achieved by enforcing assignment consistency between position-matched pairs over a set of prototypes (Fig.~\ref{localbranchfigure}). 
Specifically, we select $N$ position-matched pairs from the feature vectors in the output feature maps of the student and teacher using their absolute positions in original input image $x$, denoted as $\{(x_i, \hat{x}_i)\}_{i=1}^N$. 
However, since the masked and augmented images are generated by applying two random crops to the original image, there may not be any overlapping regions between the two cropped images~\cite{wang2021dense,xie2021propagate,muhtar2022index}. In other words, there is no position-matched feature vectors. To address this issue, we further restrict the size of the minimum cropped image in the random crop process to ensure that there is a certain overlap between the two cropped images.
The process for calculating the absolute positions is shown in Fig.~\ref{position}. Suppose $(i_2,j_2,w_2,h_2)$ representing the (left, top, height, width) positional parameters of the input cropped image, with the model output sizes being $H\times W$. 
The absolute position $(l_1, l_2)$ of the feature vector at position $(u,v)$ in the feature map can be calculated as:

 \begin{equation}
     l_1 = i_2 + \frac{w_2}{2W} + \frac{w_{2}}{W}(u-1),
 \end{equation}
 \begin{equation}
     l_2 = j_2 + \frac{h_2}{2H} + \frac{h_{2}}{H}(v-1).
 \end{equation}
We calculate the Euclidean distance of each feature vector in the two feature maps using their absolute position, and the top $N$ closest feature vectors are chosen as the position-matched pairs $\{(x_i, \hat{x}_i)\}_{i=1}^N$. 
These matched pairs are then projected into another feature space and mapped to a set of learnable prototypes to obtain their respective similarities to these prototypes. 
During the training process, these prototypes are gradually learned and model the semantic information of the various objects in the dataset, allowing the model to learn more advanced local object-level semantics than directly aligning the matched pairs using the contrastive loss~\cite{wang2021dense,muhtar2022index,li2022global}. Additionally, this approach does not require different matched pairs to correspond to different objects, thereby avoiding the false negatives problem introduced by~\cite{muhtar2022index}.
Finally, the student is trained to minimize the difference in the distributions over these similarities between the position-matched pairs. Precisely, let $C \in \mathbb{R}^{K \times d}$ denote the $K$ learnalbe prototypes with dimension $d$. The similarity distributions $p_i$ and ${q}_i$ between $x_i$ and $\hat{x}_i$ with respect to C can be expressed as:

\begin{equation}
    p_i = Softmax\left(\frac{1}{\tau_s}\langle x_i, C\rangle\right)\label{propobpilitystudent},
\end{equation}
\begin{equation}
    {q}_i = Softmax\left(\frac{1}{\tau_t}\langle\hat{x}_i, C\rangle\right)\label{propobpilityteacher},
\end{equation}
where $\langle\cdot,\cdot\rangle$ denotes the cosine similarity and $\tau_s,\tau_t$ are temperature parameters that control the sharpness of the distribution~\cite{caron2021emerging}. Finally, we minimize the cross entropy loss between $p_i$ and $q_i$:
\begin{equation}
    \mathcal{L}_{local}=\frac{1}{N}\sum_{i=1}^{N}-p_ilogq_i.
\end{equation}

The total loss of CMID is then define as:
\begin{equation}
    \mathcal{L}=\lambda_1(\mathcal{L}_ {spat} + \mathcal{L}_{freq})+\lambda_2\mathcal{L}_{NCE}+\lambda_3\mathcal{L}_{local},
\end{equation}
where $\lambda_1, \lambda_2, \lambda_3$ act as the weight to balance the three branches.

\section{Experiment}

\subsection{Implementation Detail}
We use ResNet-50~\cite{he2016deep} or Swin Transformer-Base (Swin-B)~\cite{liu2021swin} as the encoder of the student and teacher. The MIM head in the MIM branch is a 1 $\times$ 1 convolution layer, while the global projector is a one-hidden layer MLP and the local projector is a MLP with two hidden layers.  All EMA momentum rates are set to 0.996 and gradually increased to 1 by the cosine annealing scheduler during training.  The infoNCE loss temperature $\tau$ in Eq.~\ref{infonce} and the size of queue are set to 0.2 and 65536 following the best practice of MoCo~\cite{chen2020improved}. The values of $\tau_s$ and $\tau_t$ in Eq.~\ref{propobpilitystudent} and Eq.~\ref{propobpilityteacher} are set to 0.2 and 0.07, respectively, as suggested in ~\cite{caron2021emerging}. 
Based on the results of ablation experiments, we set the mask ratio, number of prototypes $K$, and the number of selected position-matched pairs $N$ to 0.6, 2048, and 20, respectively.
Finally, the weight of each branch $(\lambda_1, \lambda_2, \lambda_3)$ is set to 1 based on the results of our ablation experiments.

\subsection{Ablation Experiment}\label{ablationstudy}

\subsubsection{Dataset} The ISPRS Potsdam dataset\footnote{https://www.isprs.org/education/benchmarks/UrbanSemLab/2d-sem-label-potsdam.aspx} is used for pre-training and fine-tuning in all ablation experiments. The Potsdam dataset consists of 38 tiles of the same size of 6,000 $\times$ 6,000 pixel with a spatial resolution of 0.5m. The dataset has been manually labeled into six categories: low vegetation, tree, building, impervious surface, car, and clutter. We crop these 38 tiles of images to 256 $\times$ 256 pixel, yielding a total of 21,888 images for pre-training. For fine-tuning, we crop these images to 512 $\times$ 512 pixel and follow the default train-test split strategy provided by mmsegmentation~\cite{mmseg2020}.

\subsubsection{Experimental Setting} We choose semantic segmentation tasks to evaluate the influence of different components with different parameters on the pre-training performance of CMID, as semantic segmentation requires the pre-trained model to have the ability to extract both global and local representations~\cite{muhtar2022index,li2022global}. 
\paragraph{Pre-train Setting}
The ResNet-50~\cite{he2016deep} is used as the encoder and is pre-trained for 100 epochs on the Potsdam dataset without labels. 
The Adan~\cite{xie2022adan} optimizer is used, with a batch size of 64, an initial learning rate of 0.003125, and a weight decay of 0.02. 
The learning rate follows a cosine annealing schedule, starting from 0 with 5 warmup epochs and with the final value of 0.000001.

\paragraph{Fine-tune Setting} The fine-tuning on the segmentation tasks is implemented using mmsegmentation~\cite{mmseg2020}, with UperNet~\cite{xiao2018unified} serving as the segmentation decoder. The entire model is trained for 50 epochs with a batch size of 8 using the SGD optimizer with learning rate  of 0.01, weight decay of 0.0001 and  momentum of 0.9. The learning rate is optimized by the polynomial scheduler during the fine-tuning process. 

\subsubsection{Results}
All the reported results are the averages of two fine-tuning runs.
\paragraph{Three Branches}
We report the impact of adding or removing different branches on the CMID in Table~\ref{w/othreebranch}. The results show that the best performance is achieved by combining all three branches (row 1 and row 2). Using 
MoCo~\cite{chen2020improved} in the global branch leads to better results than using the BYOL~\cite{grill2020bootstrap} style CL method. Comparing the use of only the MIM branch (row 4) versus only the CL branches (row 5), we find that the CL performs better, further demonstrating that the MIM method is not well suited for CNN architectures. Combining the MIM and local branches (row 7) gives result that surpasses those of single or double branch configurations, indicating that the local branch can effectively compensate for the semantic incompleteness of the MIM branch. CMID collapses when only the global branch is available, as we are not using shuffleBN like the standard MoCo~\cite{chen2020improved} method, which causes the model to learn toward a cheat solution (row 8). Furthermore, the results of ablation experiments on three branches with different weights show that the best performance is achieved when the three branches are balanced (Table~\ref{branchweight}).

\begin{figure}[t]
    \centering
    \includegraphics[width=1\linewidth]{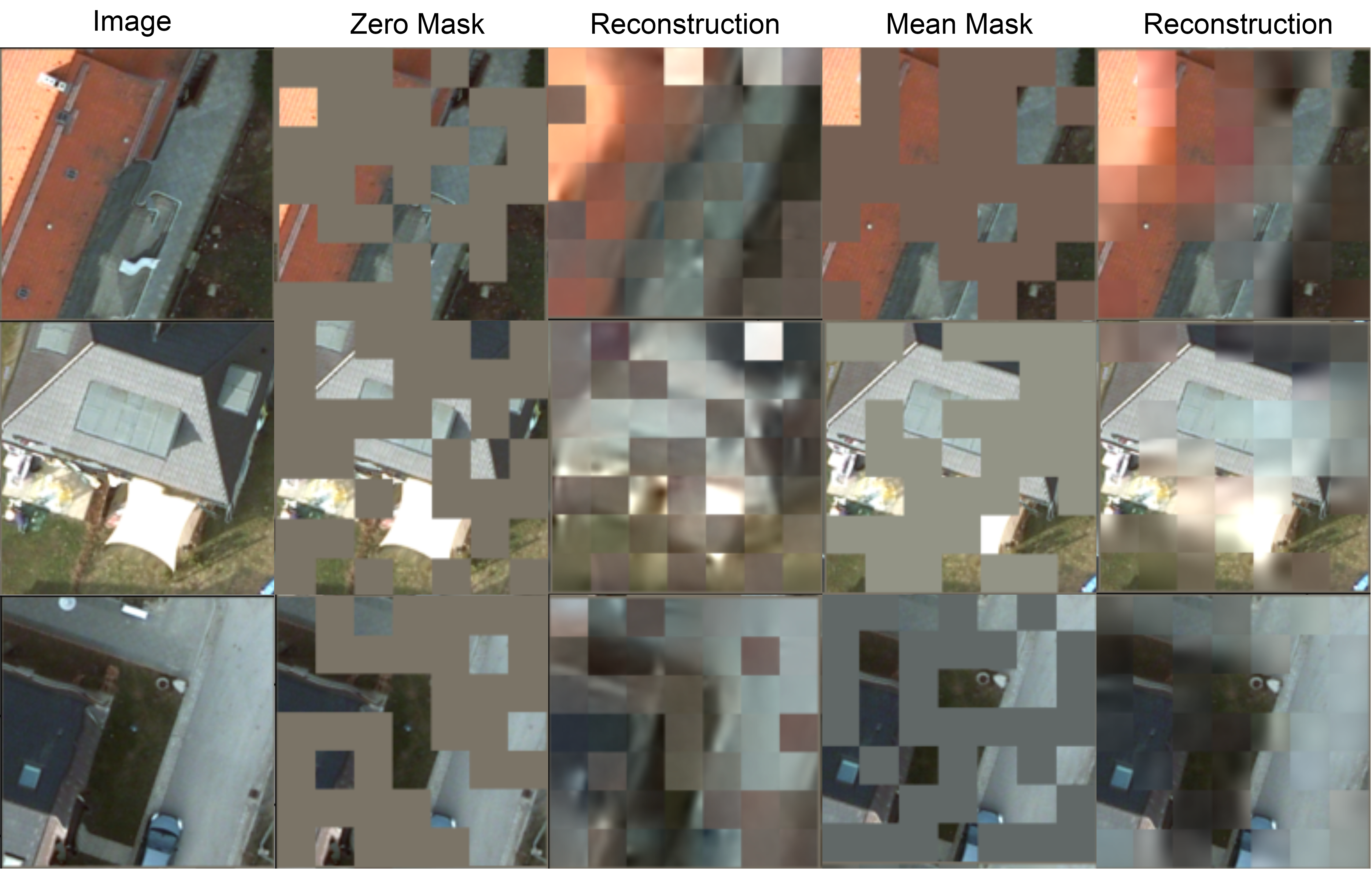}
    \caption{Reconstruction results for different masking strategies. The results show that using mean RGB value to mask image and adding the mask token to patch embedding leads to better reconstruction results, demonstrating that the proposed mask strategy can provide informative hints for the reconstruction task.}
    \label{maskvisual}
\vspace{-0.2cm}
\end{figure}

\begin{table}[t]
\caption{Ablation results for three branches.}
\vspace{-0.2cm}
\label{w/othreebranch}
\resizebox{\columnwidth}{!}{
\begin{tabular}{lcccccc}
Method & MIM & Global Branch & Local Branch & mIoU & OA & mF1 \\
\toprule
\rowcolor{cyan!50}
CMID-MoCo & \cmark                     & \cmark                     & \cmark  & 86.97 & 92.75 & 92.89 \\
CMID-BYOL & \cmark                     & \cmark                     & \cmark  & 86.33 & 92.40 & 92.51 \\
          & \cmark                     & \cmark                     & \colorbox{cyan!50}{\makebox(2,2){\xmark}} & 86.11 & 92.24 & 92.39 \\
          & \cmark                     & \colorbox{cyan!50}{\makebox(2,2){\xmark}} & \colorbox{cyan!50}{\makebox(2,2){\xmark}} & 84.54 & 91.21 & 91.37 \\
          & \colorbox{cyan!50}{\makebox(2,2){\xmark}} & \cmark                     & \cmark  & 86.06            & 92.2  & 92.34 \\
          & \colorbox{cyan!50}{\makebox(2,2){\xmark}} & \colorbox{cyan!50}{\makebox(2,2){\xmark}} & \cmark  & 85.75            & 92.02 & 92.18 \\
          & \cmark                     & \colorbox{cyan!50}{\makebox(2,2){\xmark}} & \cmark  & 86.16            & 92.26 & 92.41 \\
          & \colorbox{cyan!50}{\makebox(2,2){\xmark}} & \cmark                     & \colorbox{cyan!50}{\makebox(2,2){\xmark}} & 78.04 & 87.61 & 87.21 \\
\toprule
\end{tabular}
}
\end{table}
\begin{table}[t]
\caption{Ablation results for different branch weights.}
\vspace{-0.2cm}
\label{branchweight}
\resizebox{\columnwidth}{!}{
\begin{tabular}{cccccc}
Global Branch ($\lambda_2$)& Local Branch ($\lambda_3$)  & MIM Branch ($\lambda_1$) & mIoU & OA & mF1  \\
\toprule
\rowcolor{cyan!50}
1             &      1        &     1      & 86.97 & 92.75 & 92.89 \\
1             &     0.5       &     1      & 86.13 & 92.23 & 92.38 \\
0.5           &      1        &     1      & 86.34 & 92.38 & 92.53 \\
1             &      1        &    0.5     & 86.34 & 92.34 & 92.51 \\
\bottomrule
\end{tabular}
}
\end{table}

\paragraph{Mask Strategy}
We show the effects of using different mask strategies in Table~\ref{maskstategy}. The results show that the proposed mask strategy, which fills the mask patches with mean RGB value and adds the mask token to patch embedding, yields the best performance. This demonstrates that the proposed mask strategy results in better interaction between CL and MIM. In addition, the inclusion of reconstructions in the frequency domain further enhances the performance of the pre-trained model. Moreover, the visualization results show that the proposed masking strategy achieves better reconstruction results (Fig.~\ref{maskvisual}), further demonstrating that the proposed masking strategy can reduce the difficulty of the reconstruction task and enable the model to learn local spatial perceptible representations more efficiently.
\begin{table}[t!]
\caption{Ablation results for different mask strategies and frequency domain reconstruction.}
\centering
\vspace{-0.2cm}
\label{maskstategy}
\begin{tabular}{ccccc}
FFT        & Mask Strategy  & mIoU & OA     & mF1 \\
\toprule
\rowcolor{cyan!50}
\cmark      &  mean + add        & 86.97 & 92.75 &  92.89  \\
            &  mean + add        & 86.57 & 92.53 &  92.65  \\
\cmark     &  zero + replace     & 85.52 & 91.82 &  92.03  \\

\bottomrule
\end{tabular}
\end{table}

\begin{table}[t!]
\caption{Ablation results for different mask ratios.}
\centering
\vspace{-0.2cm}
\label{maskratio}
\begin{tabular}{ccccc}
Mask Ratio & 0.30 & 0.45    &  0.60  & 0.75 \\
\toprule
mIoU       & 86.33 & 86.35  & \textbf{86.97} & 86.68 \\ 
\bottomrule
\end{tabular}
\end{table}
\paragraph{Mask Ratio}
We conduct experiments with various mask ratios and find that a high mask ratio enables the model to learn more robust representations, leading to improved performance on downstream tasks. However, it is important to not set the mask ratio too high, as it can negatively impact model performance (Table~\ref{maskratio}). Therefore, we set the default mask ratio to 0.6.

\begin{table}[t!]
\centering
\caption{Ablation results for different minimum cropped image scales and the number of position-matched pairs. mIoU after fine-tuning is reported for each setting.}
\vspace{-0.2cm}
\label{numofmatched}
\begin{tabular*}{.7\columnwidth}{@{\extracolsep{\fill}}lccc@{}}
\toprule
$scale_{min}$ & \multicolumn{3}{c}{$N$} \\ \midrule
      & 15    & 20    & 25    \\ \midrule
0.2   & \textbf{86.52} & 86.44 & 86.05 \\
0.3   & \textbf{86.68} & 86.65 & 86.57 \\
0.4   & 86.29 & \textbf{86.69} & 86.61 \\
0.5   & 86.20 & \textbf{86.97} & 86.42 \\
0.6   & 85.94 & \textbf{86.76} & 86.19 \\ \bottomrule
\end{tabular*}%
\end{table}

\begin{table}[t!]
\centering
\caption{Ablations results for different numbers of prototypes.}
\vspace{-0.2cm}
\label{numofprototypes}
\begin{tabular}{lcccc}
\toprule
$K$        & 512    & 1024    & 2048  & 4096   \\
mIoU       & 86.43  & 86.81   & \textbf{86.97} & 86.69 \\
\bottomrule
\end{tabular}
\vspace{-0.4cm}
\end{table}

\paragraph{Number of Matched Pairs}
In this set of ablation experiments, we attempt to identify the optimal minimum cropped size and number of position-matched pairs $N$ that would yield the best performance. The results are shown in Table~\ref{numofmatched}.
$scale_{min}$ represents the ratio between the minimum size of the crop and the original size of the image.
As the scale increases, the optimal number of matched samples tends to increase, which supports the notion that aligning more local semantics can improve model performance when there is a large overlap between two cropped images. However, beyond a certain scale, model performance begins to decline. This is likely due to a decrease in representation discriminability in the  global branch as the difference between the two views becomes less different when the scale is too large~\cite{chen2020simple}. Finally, we selected the $scale_{min}$ of 0.5 and the $N$ of 20 as the default CMID settings.

\paragraph{Number of Prototypes}
We evaluate the inﬂuence of the number of prototypes $K$ used in CMID in Table~\ref{numofprototypes}. We discovered that the local branch is more robust to the number of prototypes because the results do not vary significantly. As such, we choose 2048 as the default number of prototypes.

\subsection{Pre-training}
\subsubsection{Dataset}
We use the MillionAID dataset~\cite{Long2021DiRS} for CMID pre-training. MillionAID is a massive dataset comprising 1,000,848 non-overlapping RS images. The images in MillionAID are gathered from Google Earth and captured by various sensors with varying resolutions. On average, the image sizes in the MillionAID dataset range from 110 $\times$ 110 pixel to 31,672 $\times$ 31,672 pixel. We utilize all the images in the MillionAID dataset for pre-training and resize them to 224 $\times$ 224 pixel for network input.
\subsubsection{Pre-training Setting}
We pre-train the CMID network with ResNet-50 or Swin-B as the backbone for 200 epochs in the MillionAID dataset without any label. 
The Adan optimizer~\cite{xie2022adan} is employed to optimize the model, with a learning rate of 0.0088 and a batch size of 512 for ResNet-50 and 0.002, 256 for Swin-B. We utilize the cosine learning rate scheduler with the first 15 epochs for warm-up. The weight decay is set to 0.02. Pre-training is conducted using 4 NVIDIA V100 GPUs.
For fair and holistic comparison between CMID and other SSL methods, we employed the following two comparative strategies.
\begin{itemize}
    \item [(a)] We download the open source pre-trained weights of 8 state-of-the-art SSL methods, including BYOL~\cite{grill2020bootstrap}, SwAV~\cite{caron2020unsupervised}, Barlow-Twins~\cite{zbontar2021barlow}, and MoCo-v2~\cite{chen2020improved} in the field of computer vision and SeCo~\cite{manas2021seasonal}, ResNet-50-SEN12MS~\cite{scheibenreif2022self}, ViTAE-B-RVSA~\cite{wang2022advancing}, and ViT-B-RVSA~\cite{wang2022advancing} in the field of RS. 
    We then fine-tuned them in the same settings to evaluate their performance on various RS downstream tasks  (Sec.~\ref{sec:sceneclassification}-Sec.~\ref{sec:cd}). 
    We consider that comparing with official open source pre-trained models could be more convincing than re-implementing them, as the official pre-training models of various SSL methods are highly tuned and optimized.
    \item [(b)] We pre-trained the models using different SSL methods (CMID, MAE~\cite{he2022masked}, SimMIM~\cite{xie2022simmim}, BYOL~\cite{grill2020bootstrap}, Barlow-Twins~\cite{zbontar2021barlow}, MoCo-v2~\cite{chen2020improved}) under the same pre-training settings, and subsequently evaluated their performance on RS downstream tasks. This was done to eliminate any potential influence arising from different pre-training datasets and evaluate the scalability of different SSL methods on small datasets (Sec.~\ref{sec:scale}).
\end{itemize}

\subsection{Scene Classification}\label{sec:sceneclassification}
\subsubsection{UC Merced Land Use Dataset} The UC Merced Land Use (UCM) dataset~\cite{yang2010bag} contains 2,100 images with a resolution of 0.3 m. All the images in UCM dataset have the size of 256 $\times$ 256 pixel. The 2,100 images equally belong to 21 categories. Thus, each category has 100 images.

\begin{figure}[t]
    \centering
    \includegraphics[width=1\linewidth]{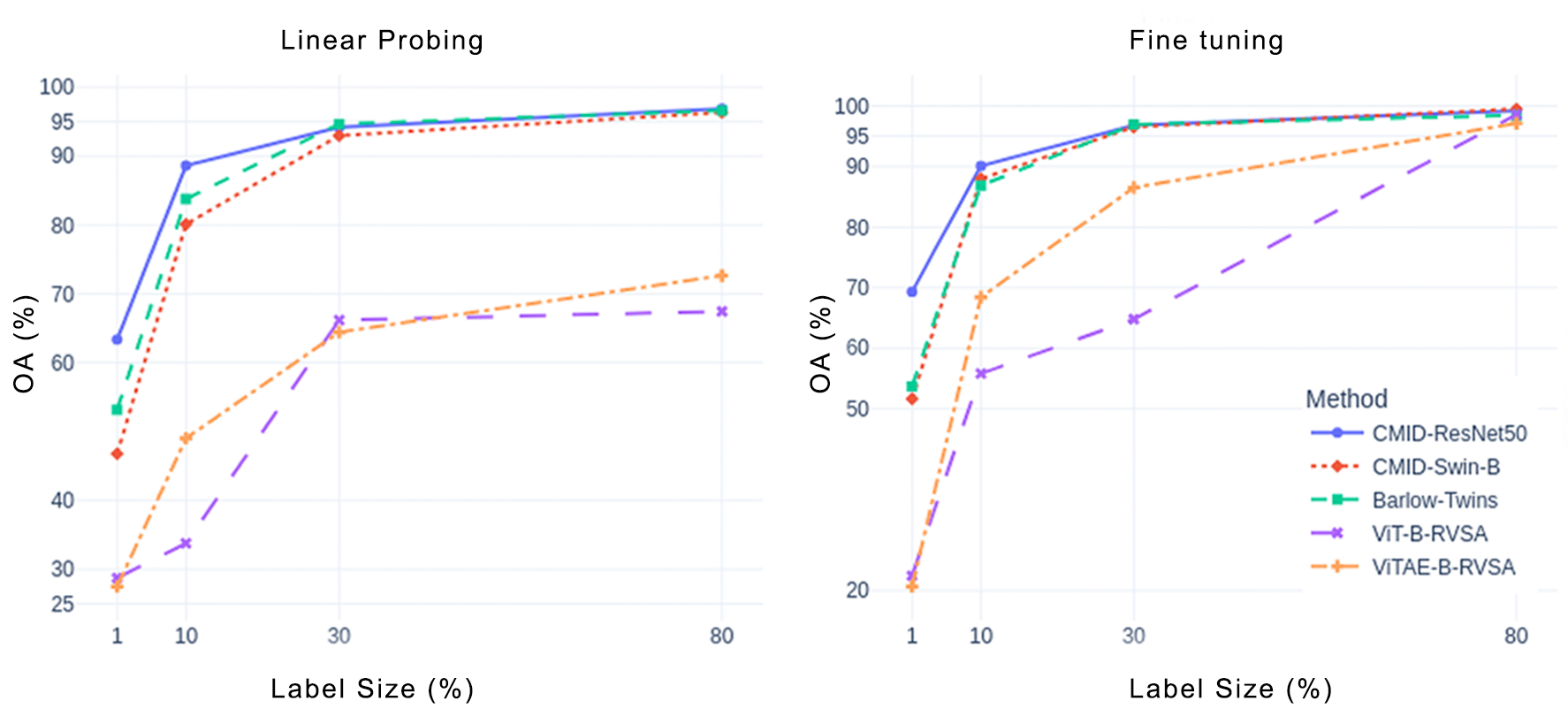}
    \caption{Evaluation on the UCM classification task with different labels size. For comparison of various ResNet-50 based pre-trained models, only the results of the comparison with Barlow-Twins are shown for ease of viewing. Barlow-Twins outperformed the other ResNet-50 based pre-trained models, except for CMID, in the classification task.}
    \label{clsresultfigure}
\vspace{-0.2cm}
\end{figure}
\begin{table}[t]
\centering
\caption{Performance of various SSL methods on the UCM testing set. The ratio of training set to testing set is 8:2.}
\vspace{-0.2cm}
\label{clsresult}
\resizebox{\columnwidth}{!}{%
\begin{tabular}{lccc|c|c|}
\hline
\multirow{2}{*}{Method} & \multirow{2}{*}{Backbone} & \multirow{2}{*}{Pre-train Dataset} & \multirow{2}{*}{Pre-train Epoch} & Linear Probe & Fine tune \\ \cline{5-6} 
                  &              &            &      & OA    & OA    \\ \hline
ImageNet Sup.     & ResNet-50    & ImageNet   & -    & 94.79 & 97.92 \\
BYOL              & ResNet-50    & ImageNet   & 200  & 93.23 & 99.22 \\
Barlow-Twins      & ResNet-50    & ImageNet   & 300  & 96.61 & 99.16 \\
MoCo-v2           & ResNet-50    & ImageNet   & 200  & 88.80 & 97.92 \\
SwAV              & ResNet-50    & ImageNet   & 200  & 94.79 & 98.96 \\
SeCo              & ResNet-50    & SeCo-1M    & 200  & 90.36 & 97.66 \\
ResNet-50-SEN12MS & ResNet-50    & SEN12MS    & 200  & 73.18 & 96.88 \\
MAE               & ViT-B-RVSA   & MillionAID & 1600 & 67.45 & 98.56 \\
MAE               & ViTAE-B-RVSA & MillionAID & 1600 & 72.66 & 97.12 \\
\rowcolor{cyan!50}
CMID              & ResNet-50    & MillionAID & 200  & 96.88 & 99.22 \\
\rowcolor{cyan!50}
CMID              & Swin-B       & MillionAID & 200  & 96.35 & 99.48 \\ \hline
\end{tabular}%
}
\end{table}
\subsubsection{Experimental Setting}
We divide the UCM dataset into a training set and a testing set randomly according to a specific ratio (1:99, 1:9, 3:7, 8:2). We use the Adan optimizer~\cite{xie2022adan} with a cosine learning rate scheduler and train for 200 epochs. The batch size is set to 64, the base learning rate is set to 0.003125, and the weight decay is set to 0.02. We also employ warmup for 10 epochs. We follow the same data augmentation setting as in~\cite{wang2022advancing}, including RandAug~\cite{cubukpractical}, Mixup~\cite{zhang2017mixup}, Cutmix~\cite{yun2019cutmix}, label smoothing~\cite{szegedy2016rethinking}, and random erasing~\cite{zhong2020random}. During the fine-tuning process, both the backbone and the classification head are updated, but in linear probing, only the classification head is updated. Overall accuracy (OA) is used as the evaluation metric in this downstream task.
\begin{figure*}[h!t!]
    \centering
    \includegraphics[width=1.0\linewidth]{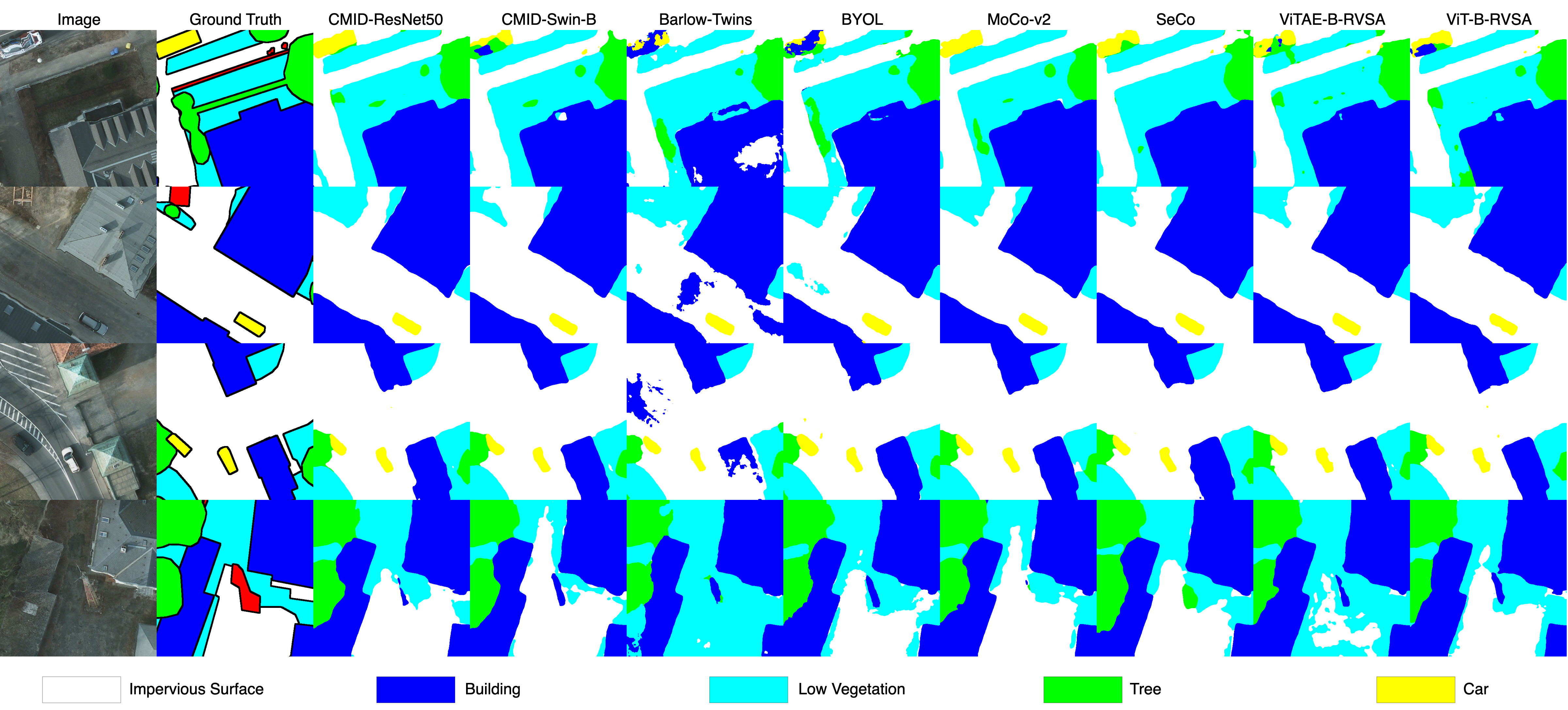}
    \caption{Visualization results of different methods on the Potsdam dataset. In inference, we disregard the background class, which is designated as red in the ground truth labels.}
    \label{semanticvisualize}
\end{figure*}

\subsubsection{Results} The experimental results in Table~\ref{clsresult} show that all pre-trained models achieve promising results when fine-tuning with 80\% of the labels. 
Our CMID pre-trained achieves the highest accuracy. 
Previous studies have shown that the models pre-trained using MIM method excel in TL with full model fine-tuning, but perform poorly in linear probing~\cite{he2022masked,tao2022siamese,zhou2022mimco}. However, the CMID pre-trained models also perform well in linear probing and outperform all other SSL methods, further demonstrating the effectiveness of integrating the MIM and CL methods. 
It is worth noting that ViTs (Swin-B, ViTAE-B-RVSA, and ViT-B-RVSA) do not perform as well as the CNN network in linear probing, which can be attributed to the different inductive biases of the two architectures. 
Additionally, we show in Fig.~\ref{clsresultfigure} the results of fine-tuning and linear probing with different label sizes. CMID pre-trained models show strong performance even when only a small number of labels are available, indicating the robustness of the learned representations during pre-training. 
Especially in the one-shot scenario (corresponding to the 1\% labels, with one sample per category), CMID substantially outperforms its ResNet-50 or ViTs counterparts. 
Fig.~\ref{clsresultfigure} also shows that, although the CMID pre-trained model based on Swin-B performs best with sufficient labels (TOP 1 with 80\% labels), it does not perform as well as the ResNet-50 series with limited labels, which we believe is due to the data-hungry nature of ViTs~\cite{dosovitskiy2020image}.

\begin{figure}[t]
    \centering
    \includegraphics[width=1\linewidth]{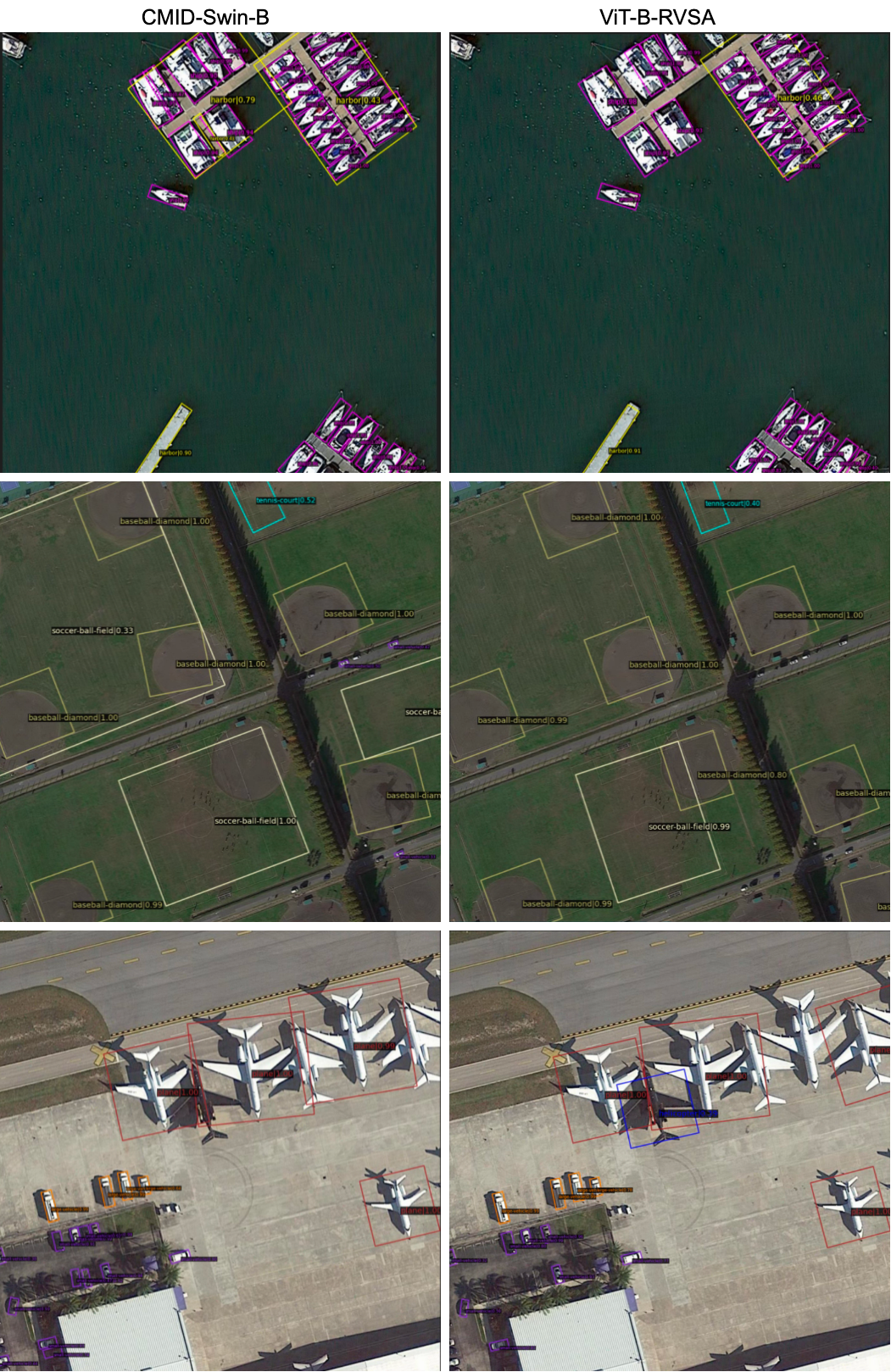}
    \vspace{-0.7cm}
    \caption{Visualization of detection results on DOTA testing set. The results show that the proposed CMID-Swin-B pre-trained model demonstrates comparable detection results to those of the ViT-B-RVSA.}
    \label{fig:detection}
\vspace{-0.5cm}
\end{figure}
\subsection{Semantic Segmentation}\label{semanticsegmentationsection}
\subsubsection{Dataset} We evaluate the semantic segmentation performance of the pre-trained models on the Potsdam and Vaihingen\footnote{https://www.isprs.org/education/benchmarks/UrbanSemLab/2d-sem-label-vaihingen.aspx} datasets. Potsdam has been described in Sec.~\ref{ablationstudy}. The Vaihingen dataset contains 33 tiles with a size of 2,494 $\times$ 2,064 pixel and a spatial resolution of 0.9 m. We crop both datasets into patches of 512 $\times$ 512 pixel with a stride of 256 pixel and follow the default train-test split strategy provided by mmsegmentation~\cite{mmseg2020}.
\begin{table}[th!]
\centering
\caption{Semantic segmentation results of different SSL pre-trained models on the Potsdam and Vaihingen datasets.}
\vspace{-0.2cm}
\label{semanticsegmentationresult}
\resizebox{\columnwidth}{!}{%
\begin{tabular}{lccc|cll|cll|}
\hline
\multirow{2}{*}{Method} &
  \multirow{2}{*}{Backbone} &
  \multirow{2}{*}{Pre-train Dataset} &
  \multirow{2}{*}{Pre-train Epoch} &
  \multicolumn{3}{c|}{Potsdam} &
  \multicolumn{3}{c|}{Vaigingen} \\ \cline{5-10} 
                  &              &            &      & OA    & mIoU  & mF1   & OA    & mIoU  & mF1   \\ \hline
ImageNet Sup.     & ResNet-50    & ImageNet   & -    & 92.00 & 85.69 & 92.11 & 90.21 & 78.53 & 87.65 \\
BYOL              & ResNet-50    & ImageNet   & 200  & 91.86 & 85.54 & 92.07 & 87.69 & 73.17 & 83.97 \\
Barlow-Twins      & ResNet-50    & ImageNet   & 300  & 90.32 & 83.16 & 90.67 & 86.58 & 71.86 & 82.02 \\
MoCo-v2           & ResNet-50    & ImageNet   & 200  & 92.74 & 87.02 & 92.93 & 90.26 & 79.16 & 88.10 \\
SwAV              & ResNet-50    & ImageNet   & 200  & 92.02 & 85.74 & 92.20 & 87.95 & 73.76 & 84.40 \\
SeCo              & ResNet-50    & SeCo-1M    & 200  & 92.04 & 85.82 & 92.22 & 90.14 & 78.59 & 87.72 \\
ResNet-50-SEN12MS & ResNet-50    & SEN12MS    & 200  & 90.50 & 83.17 & 90.62 & 88.62 & 73.99 & 84.41 \\
MAE               & ViT-B-RVSA   & MillionAID & 1600 & 92.40 & 86.37 & 92.55 & 89.54 & 77.29 & 86.87 \\
MAE               & ViTAE-B-RVSA & MillionAID & 1600 & 92.62 & 86.61 & 92.68 & 89.79 & 78.17 & 87.49 \\
\rowcolor{cyan!50}
CMID              & ResNet-50    & MillionAID & 200  & 92.94 & 87.35 & 93.11 & 90.23 & 79.44 & 88.30 \\
\rowcolor{cyan!50}
CMID              & Swin-B       & MillionAID & 200  & 93.52 & 88.36 & 93.70 & 90.50 & 80.01 & 88.68       \\ \hline
\end{tabular}%
}
\end{table}
\begin{table}[t]
\centering
\caption{Flops and parameters size of different backbone. B,C,H,W represent the batch size, number of channels and image width and height respectively.}
\vspace{-0.2cm}
\label{param}
\begin{threeparttable}
\begin{tabular}{lccc}
\hline
Backbone & \begin{tabular}[c]{@{}c@{}}Input Size\\ (B, C, H, W)\end{tabular} & FLOPs (G) & Param. (M) \\ \hline
ResNet-50    & (1,3,224,224) & 8.26  & 23.55 \\
ViT-B-RVSA   & (1,3,224,224) & 33.56 & 86.00 \\
ViTAE-B-RVSA & (1,3,224,224) & 26.33 & 67.54 \\
Swin-B       & (1,3,224,224) & 10.61 & 30.84 \\ \hline
\end{tabular}%
  \begin{tablenotes}
    \scriptsize
    \item[1] The results are evaluated for each backbone in NVIDIA 3080 GPU using the THOP library.  
  \end{tablenotes} 
\end{threeparttable}
\end{table}
\begin{table*}[t!]
\centering
\caption{Results of different SSL pre-trained models on the testing set of the DOTA dataset.}
\vspace{-0.2cm}
\label{detection}
\begin{threeparttable}
\resizebox{\linewidth}{!}{
\begin{tabular}{lccc|ccccccccccccccc|c}
\hline
Method &
  Backbone &
  Pre-train Dataset &
  Pre-train Epoch &
  PL &
  BD &
  BR &
  GTF &
  SV &
  LV &
  SH &
  TC &
  BC &
  ST &
  SBF &
  RA &
  HA &
  SP &
  HC &
  mAP \\ \hline
ImageNet Sup. &
  ResNet-50 &
  ImageNet &
  - &
  89.12 &
  83.08 &
  52.42 &
  72.99 &
  78.22 &
  81.56 &
  87.90 &
  90.90 &
  85.81 &
  85.01 &
  62.74 &
  65.22 &
  72.78 &
  68.80 &
  \multicolumn{1}{l|}{57.41} &
  75.62 \\
BYOL &
  ResNet-50 &
  ImageNet &
  200 &
  89.35 &
  79.66 &
  49.66 &
  71.16 &
  77.37 &
  81.17 &
  87.82 &
  90.87 &
  85.93 &
  85.27 &
  49.78 &
  63.03 &
  66.48 &
  66.21 &
  \multicolumn{1}{l|}{61.14} &
  73.62 \\
Barlow-Twins &
  ResNet-50 &
  ImageNet &
  300 &
  87.72 &
  70.96 &
  42.29 &
  64.56 &
  74.64 &
  75.45 &
  85.65 &
  90.83 &
  82.25 &
  79.70 &
  38.01 &
  57.81 &
  53.57 &
  64.94 &
  \multicolumn{1}{l|}{44.76} &
  67.54 \\
MoCo-v2 &
  ResNet-50 &
  ImageNet &
  200 &
  89.27 &
  78.29 &
  51.21 &
  71.67 &
  78.40 &
  81.32 &
  87.92 &
  90.89 &
  84.34 &
  84.88 &
  56.30 &
  60.91 &
  67.39 &
  65.32 &
  \multicolumn{1}{l|}{50.72} &
  73.25 \\
SwAV &
  ResNet-50 &
  ImageNet &
  200 &
  89.56 &
  78.69 &
  48.74 &
  69.93 &
  77.10 &
  80.56 &
  87.64 &
  90.90 &
  86.40 &
  84.43 &
  52.42 &
  61.87 &
  64.52 &
  63.65 &
  \multicolumn{1}{l|}{63.20} &
  73.30 \\
SeCo$^{\dagger}$ &
  ResNet-50 &
  SeCo-1M &
  200 &
  88.64 &
  73.32 &
  49.94 &
  67.07 &
  76.15 &
  76.48 &
  70.02 &
  90.88 &
  79.46 &
  81.31 &
  41.31 &
  58.55 &
  65.90 &
  65.32 &
  \multicolumn{1}{l|}{49.41} &
  70.02 \\
MAE &
  ViT-B-RVSA &
  MillionAID &
  1600 &
  89.08 &
  82.34 &
  55.18 &
  72.37 &
  79.23 &
  85.01 &
  88.17 &
  90.88 &
  87.40 &
  86.81 &
  60.66 &
  68.72 &
  76.68 &
  78.35 &
  \multicolumn{1}{l|}{70.33} &
  78.08 \\
MAE &
  ViTAE-B-RVSA &
  MillionAID &
  1600 &
  89.28 &
  76.97 &
  53.85 &
  69.67 &
  79.79 &
  84.93 &
  88.31 &
  90.84 &
  85.36 &
  85.56 &
  59.15 &
  67.50 &
  77.17 &
  80.08 &
  \multicolumn{1}{l|}{65.91} &
  76.96 \\
\rowcolor{cyan!50}
CMID &
  ResNet-50 &
  MillionAID &
  200 &
  89.34 &
  84.60 &
  52.55 &
  73.69 &
  78.10 &
  82.71 &
  87.60 &
  90.88 &
  87.43 &
  85.54 &
  67.11 &
  64.97 &
  74.35 &
  71.51 &
  \multicolumn{1}{l|}{59.04} &
  76.63 \\
\rowcolor{cyan!50}
CMID &
  Swin-B &
  MillionAID &
  200 &
   89.46 &
   83.54 &
   53.51 &
   74.79 &
   78.72 &
   84.14 &
   87.58 &
   90.90 &
   86.08 &
   85.77 &
   65.97 &
   68.05 &
   74.02 &
   73.27 &
  \multicolumn{1}{l|}{64.62} &
   77.36
   \\ \hline
\end{tabular}%
}
  \begin{tablenotes}
    \scriptsize
    \item[1] PL: plane. BD: baseball diamond. BR: bridge. GTF: ground track field. SV: small vehicle. LV: large vehicle. SH: ship. TC: tennis court. BC: baseball court. ST: storage tank. \\ SBF: soccer ball field. RA: roundabout. HA: harbor. SP: swimming pool. HC: helicopter.
    \item[2] $\dagger$ represents the result from~\cite{wang2022empirical}.
  \end{tablenotes} 
  \end{threeparttable}
\end{table*}
\subsubsection{Experimental Setting} We adopt the UperNet~\cite{xiao2018unified} framework with the different SSL pre-trained models as its backbone. We fine-tune for 50 epochs with a batch size of 8. The ResNet-50 based networks are optimized using the SGD optimizer, where the initial learning rate, weight decay and momentum are set to 0.01, 0.0005 and 0.9, respectively. The learning rate is adapted by the cosine annealing scheduler. While for the Swin-B, we use the AdamW optimizer with a learning rate of 0.00006 and weight decay of 0.05. For ViT-B-RVSA and ViTAE-B-RVSA, we follow the official fine-tune settings of~\cite{wang2022advancing}.
\subsubsection{Results} Table~\ref{semanticsegmentationresult} presents the segmentation results of CMID and other SSL methods. When using either ResNet-50 or Swin-B, the CMID pre-trained model outperforms the other SSL methods in terms of OA, mean intersection over union (mIoU), and mean F1-score (mF1). 
In comparison to the ViT-B-RVSA and ViTAE-B-RVSA models, which were pre-trained for 1600 epochs, CMID demonstrates a significantly higher level of performance, highlighting the efficiency of CMID's pre-training.
Despite the ViT-B-RVSA and ViTAE-B-RVSA~\cite{wang2022advancing} are larger and more powerful than ResNet-50 (Table~\ref{param}), the CMID model, based on ResNet-50, outperforms them in the semantic segmentation task, further indicating that CMID learns more rich and useful representations during the pre-training stage.
Additionally, the visualization results in Fig.~\ref{semanticvisualize} demonstrates that CMID pre-trained models consistently exhibit superior performance.
Since semantic segmentation requires both global and local representations~\cite{li2022global,muhtar2022index}, the excellent performance of CMID in semantic segmentation demonstrates that CMID can learn representations with both global semantic separability and local spatial perceptibility during pre-training.

\subsection{Object Detection}\label{sec:objectdetection}
\subsubsection{Dataset} DOTA~\cite{xia2018dota} is a large-scale dataset for oriented bounding box (OBB) detection, with a total of 2,806 images ranging in size from 800 $\times$ 800 pixel to 4,000 $\times$ 4,000 pixel. These images contain a total of  188,282 labeled instances belonging to 15 categories, which are divided into training, validation, and testing sets with 1,411, 458, and 937 images, respectively. Following the official setting of mmrotation~\cite{zhou2022mmrotate}, we sample and crop the DOTA dataset to patches of 1,024 $\times$ 1,024 pixel with a stride of 824 pixel.
\subsubsection{Experimental Setting} We adopt the OBB detection framework ORCN~\cite{xie2021oriented} to perform TL to object detection in mmrotation~\cite{zhou2022mmrotate}. For evaluation on ResNet-50, we use the SGD optimizer with a learning rate of 0.005, weight decay of 0.0001, and a maximum gradient norm of 35.0. For Swin-B, we use AdamW optimizer with a learning rate and weight decay of 0.0001 and 0.05, respectively. For ViT-B-RVSA and ViTAE-B-RVSA, we follow the official fine-tuning hyper-parameters~\cite{wang2022advancing}. All models are trained for 12 epochs with a batch size of 2, and the learning rate is reduced by a factor of 10 after the 8th and 11th epochs. We report the mean average precision (mAP) and all evaluation results are obtained by submitting the predictions for the testing set to the online server.
\subsubsection{Results}
The results of the OBB detection are shown in Table~\ref{detection}. The detector using the CMID pre-trained ResNet-50 as its backbone surpasses other ResNet-50 based self-supervised pre-trained models by a significant margin. It is worth noting that while ViTAE-B-RVSA and ViT-RVSA~\cite{wang2022advancing} use a longer pre-training strategy and employ a  rotational window attention mechanism for RS images object detection task, our CMID-ResNet50 achieves comparable results to ViTAE-B-RVSA and the proposed CMID-Swin-B even outperforms the ViTAE-B-RVSA and yields comparable results to ViT-B-RVSA. The visualization results in Fig.~\ref{fig:detection} also suggest that the detector using the Swin-B pre-trained by CMID as the backbone achieves comparable performance with the ViT-B-RVSA. 

\begin{table}[t!]
\centering
\caption{Change detection results of different SSL pre-trained models on the CDD testing set.}
\vspace{-0.2cm}
\label{changedetection}
\begin{threeparttable}
\resizebox{\columnwidth}{!}{%
\begin{tabular}{lccc|c}
\hline
Method            & Backbone  & Pre-train Dataset & Pre-train Epoch & mF1   \\ \hline
ImageNet Sup.$^{\dagger}$     & ResNet-50 & ImageNet          & -               & 95.09 \\
BYOL              & ResNet-50 & ImageNet          & 200             & 96.30 \\
Barlow-Twins      & ResNet-50 & ImageNet          & 300             & 95.63 \\
MoCo-v2           & ResNet-50 & ImageNet          & 200             & 96.05 \\
SwAV              & ResNet-50 & ImageNet          & 200             & 95.89 \\
SeCo              & ResNet-50 & SeCo-1M           & 200             & 96.26 \\
ResNet-50-SEN12MS & ResNet-50 & SEN12MS           & 200             & 95.88 \\
ImageNet Sup.$^{\dagger}$     & ViTAEv2-S & ImageNet          & 300             & 97.02 \\
MillionAID Sup.$^{\dagger}$   & ViTAEv2-S & MillionAID        & 300             & 96.81 \\
\rowcolor{cyan!50}
CMID              & ResNet-50 & MillionAID        & 200             & 96.95 \\
\rowcolor{cyan!50}
CMID              & Swin-B    & MillionAID        & 200             &  97.11     \\ \hline
\end{tabular}%
}
  \begin{tablenotes}
    \scriptsize
    \item[1] $\dagger$ represents the result from~\cite{wang2022empirical}.
  \end{tablenotes} 
  \end{threeparttable}
\end{table}
\subsection{Change Detection}\label{sec:cd}
\subsubsection{Dataset} The CDD dataset~\cite{lebedev2018change}, which contains 11 pairs of bi-temporal images obtained from Google Earth at different seasons, has a spatial resolution that ranges from 3 to 100 cm per pixel. These images include 7 pairs of images with the size of 4,725 $\times$ 2,200 pixel and 4 images with a size of 1,900 $\times$ 1,000 pixel. We crop the images to a series of patches of 256 $\times$ 256 pixel following~\cite{ji2018fully} and generate 10,000 image pairs for training, 3,000 images pairs for validation, and 3,000 image pairs for testing.
\begin{figure}[t!]
    \centering
    \includegraphics[width=1\linewidth]{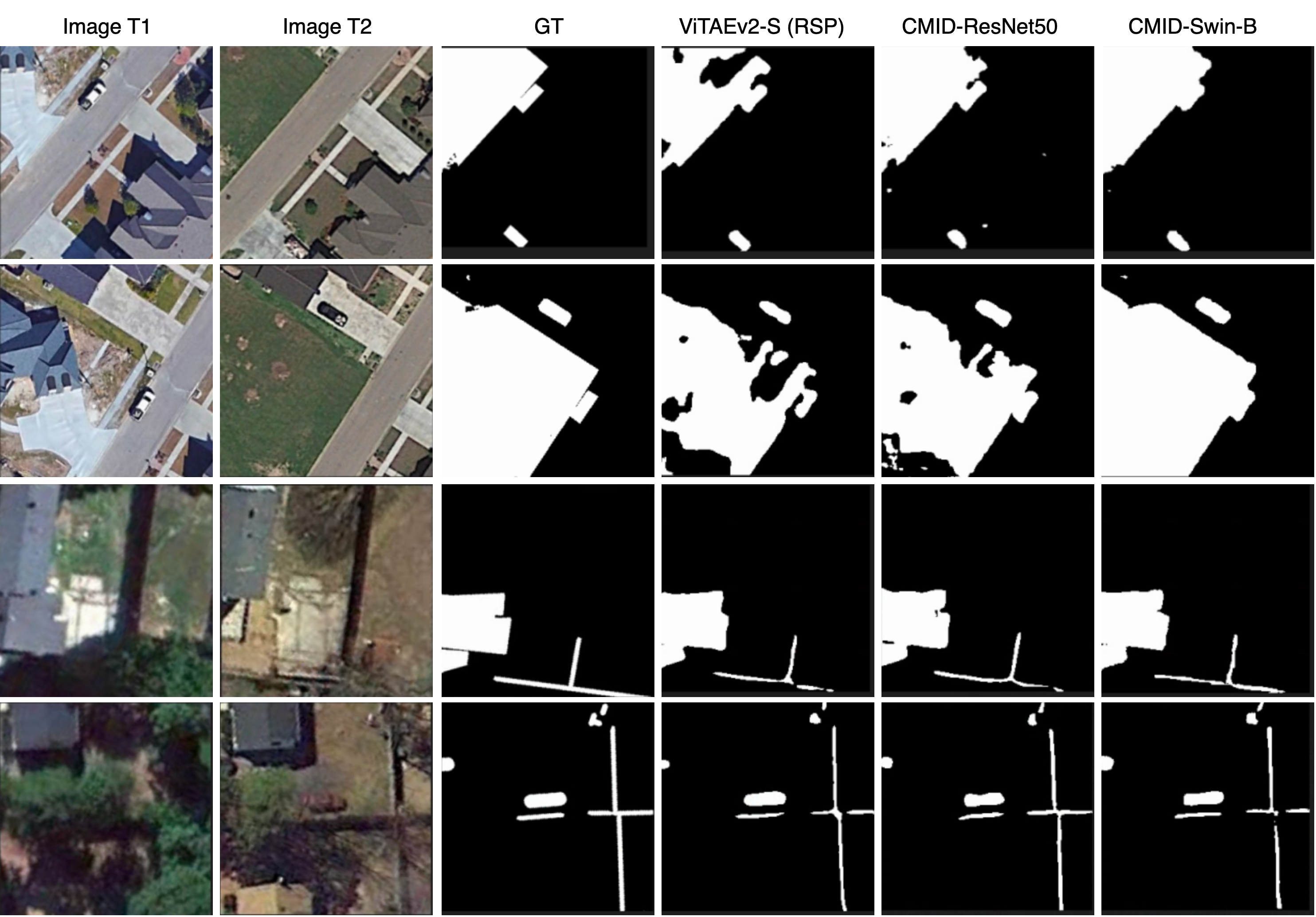}
    \vspace{-0.6cm}
    \caption{Visualization of change detection results on CDD testing set. Image T1 and image T2 are the ﬁrst second temporals of the same regions. GT refers to ground truth.}
    \label{fig:cd_result}
\vspace{-0.2cm}
\end{figure}
\begin{figure*}[t]
    \centering
    \includegraphics[width=1.0\linewidth]{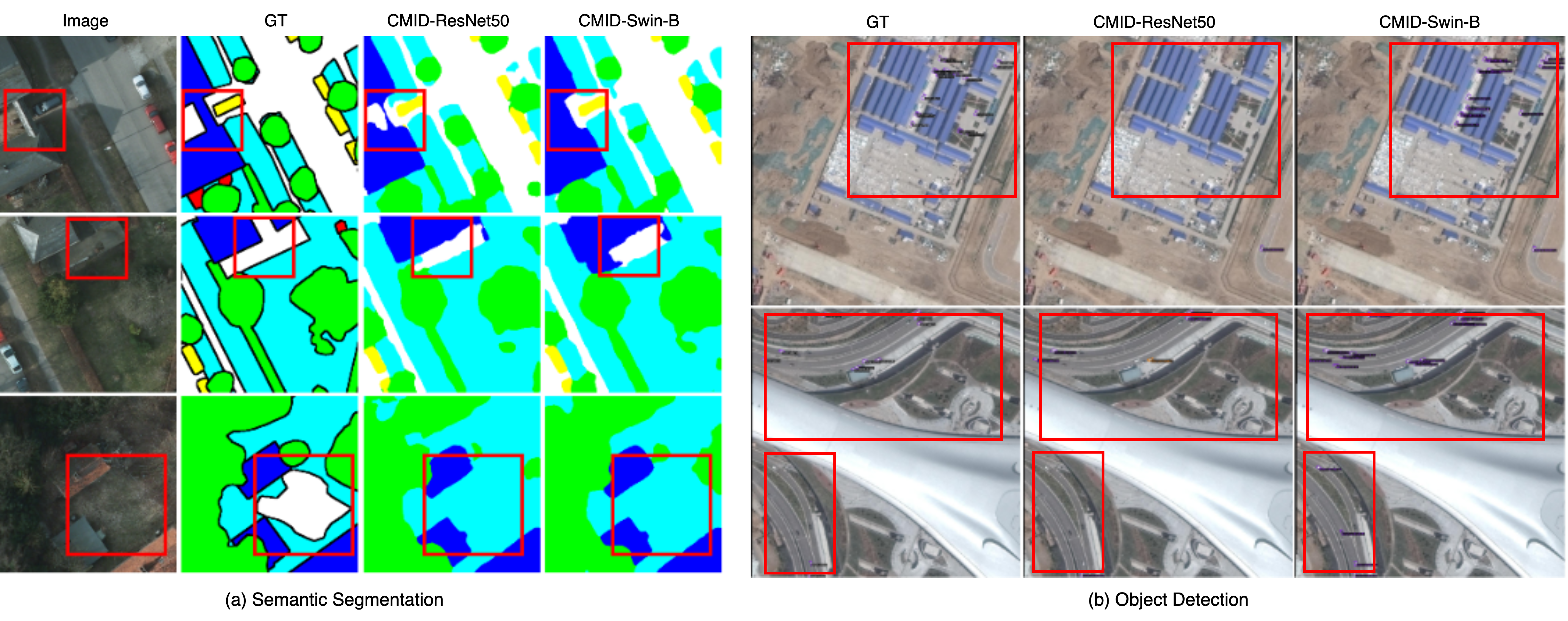}
    \vspace{-0.5cm}
    \caption{Visualization of failure cases when fine-tuning the CMID pre-trained models for semantic segmentation and object detection tasks. GT refers to ground truth.}
    \label{fig:failExample}
\end{figure*}
\subsubsection{Experimental Setting}Following ~\cite{wang2022empirical}, we adopt the BiT~\cite{chen2021remote} framework for evaluation on change detection task. 
We use the SGD optimizer with a learning rate of 0.001, a batch size of 16, and a weight decay of 0.0005 for the ResNet-50 series.
For the BiT models based on Swin-B or ViTAE, we use the AdamW optimizer with a learning rate of 0.00006 and a weight decay of 0.01.
All models are trained for 200 epochs, with the learning rate decreasing linearly during the training process. Once the fine-tuning is complete, we report the mean F1-score (mF1) on the testing set.

\subsubsection{Results}
The results of the change detection are presented in Table~\ref{changedetection}. The BiT model based on CMID-ResNet50 performs better than other ResNet50-based models and even outperforms the ViTAEv2-S model, which is pre-trained on the MillionAID dataset with supervision. 
In addition, the proposed CMID-Swin-B model demonstrates the best performance on the change detection tasks. 
The visualization results on the CDD change detection dataset (Fig.~\ref{fig:cd_result}) further indicate the capability of the CMID pre-trained models in detecting small variations in the images and making more precise predictions.
The excellent performance of CMID in the change detection task demonstrates that the representations learned by CMID are invariant enough to avoid pseudo-changes caused by imaging conditions, and because the learned representations are separable, the model pre-trained by CMID is able to detect changes in different objects within the image.

\subsection{Scalability}\label{sec:scale}
We pre-trained CMID and several other SSL methods on the relatively small Potsdam dataset under the same pre-training settings to eliminate any possible effects resulting from variations between natural and RS images and evaluate the scalability of different SSL methods. 
Afterwards, we fine-tuned these models to the Potsdam semantic segmentation task to assess the in-domain generalization ability of different SSL methods. Additionally, we fine-tuned these pre-trained models to the more challenging DOTA OBB detection task to further evaluate their cross-domain generalization ability.

\subsubsection{Experimental Setting}
The Potsdam dataset is processed in the same manner as described in Sec.~\ref{ablationstudy} during pre-training, without the use of any labels. All models are pre-trained from scratch for 400 epochs with a batch size of 64 in the Potsdam dataset. The Adan optimizer~\cite{xie2022adan} with a learning rate of 0.003125 and a weight decay of 0.02 is used for the CMID model. In addition, we implement five state-of-the-art SSL methods (MoCo~\cite{wang2022advancing}, BYOL~\cite{grill2020bootstrap}, Barlow-Twins~\cite{zbontar2021barlow}, MAE~\cite{he2022masked} , and SimMIM~\cite{xie2022simmim}) using the mmselfsup library~\cite{mmselfsup2021}, with hyperparameters taken from their official settings, except that the learning rate is scaled linearly according to the batch size.

After pre-training, all pre-trained models are evaluated on the Potsdam semantic segmentaion task and the DOTA OBB detection task using the same settings described in Sec.~\ref{semanticsegmentationsection} and Sec.~\ref{sec:objectdetection}.

\begin{table}[t!]
\caption{Potsdam semantic segmentation and DOTA OBB detection results of different SSL methods. All the models are pre-trained on the Potsdam dataset.}
\vspace{-0.2cm}
\label{tab:scalability}
\begin{tabular}{lccccc}
\hline
\multicolumn{1}{c}{\multirow{2}{*}{Method}} & \multirow{2}{*}{Backbone} & \multicolumn{3}{c}{Potsdam} & DOTA  \\ \cline{3-6} 
\multicolumn{1}{c}{}                        &                           & OA      & mIoU    & mF1     & mAP   \\ \hline
ImageNet Sup.                               & ResNet-50                 & 92.00   & 85.69   & 92.11   & 75.62 \\
BYOL                                        & ResNet-50                 & 90.86   & 83.80   & 91.00   & 59.35 \\
Barlow-Twins                                & ResNet-50                 & 88.31   & 79.47   & 88.31   & 57.20 \\
MoCo-v2                                     & ResNet-50                 & 91.51   & 84.88   & 91.65   & 66.90 \\
MAE                                         & ViT-B                     & 83.48   & 71.70   & 83.22   & 50.17 \\
SimMIM                                      & Swin-B                    & 89.37   & 81.09   & 89.33   & 64.85 \\
\rowcolor{cyan!50}
CMID                                        & ResNet-50                 & 92.74   & 87.04   & 92.81   & 72.12 \\
\rowcolor{cyan!50}
CMID                                        & Swin-B                    & 91.77   & 85.17   & 91.83   & 70.45 \\ \hline
\end{tabular}
\end{table}

\subsubsection{Results}
The results are presented in Table~\ref{tab:scalability}. Despite the fact that all SSL methods pre-trained on a small dataset yield degraded results compared to their counterparts trained on a larger dataset, CMID outperforms all other SSL methods and even achieves comparable results to models pre-trained with other SSL methods (e.g. MoCo-v2, BYOL) on large datasets. This is especially evident in the more challenging DOTA OBB detection task, where CMID pre-trained models outperform other self-supervised pre-trained models by a large margin, highlighting the superior scalability and generalization ability of CMID. It is also worth noting that ViTs do not perform as consistently as CNNs in this setting. This can be attributed to the data-hungry nature of ViTs and the findings of recent research on the relationship between model size and pre-training dataset size~\cite{li2022understanding}, which suggest that higher learning performance can be achieved with smaller models on a smaller dataset, while larger models may not perform as well as smaller ones. This is further supported by the CMID results, where the CMID-Swin-B model pre-trained on the MillionAID dataset performs better, but the CMID-Swin-B model does not perform as well as the CMID-ResNet50 model in this scalability evaluation settings.

\subsection{Failure Case}
In this subsection, we visualize the failure cases encountered when fine-tuning the CMID pre-trained models for semantic segmentation and object detection, as shown in Fig.~\ref{fig:failExample}. Our aim is to examine the limitations of CMID and offer guidance for future improvement.
\subsubsection{Semantic Segmentation} 
We visualize the semantic segmentation results on the Potsdam dataset in Fig.~\ref{fig:failExample}~(a). The pre-training and fine-tuning settings are the same as Sec.~\ref{semanticsegmentationsection}. The results show that the pre-trained CMID model is inclined to mix up the two categories when they exhibit a high level of similarity. For example, the impervious surfaces and buildings in the first and second rows share a similar color and texture, and the shadows caused by the imaging angle or lighting conditions amplify this similarity, causing the model to misinterpret the two categories. 
We firmly believe that resolving this issue necessitates an understanding of the interdependence between different categories and the implementation of constraints on this relationship to prevent any misunderstandings or mix-ups between categories.
\begin{figure}[t!]
    \centering
    \includegraphics[width=1\linewidth]{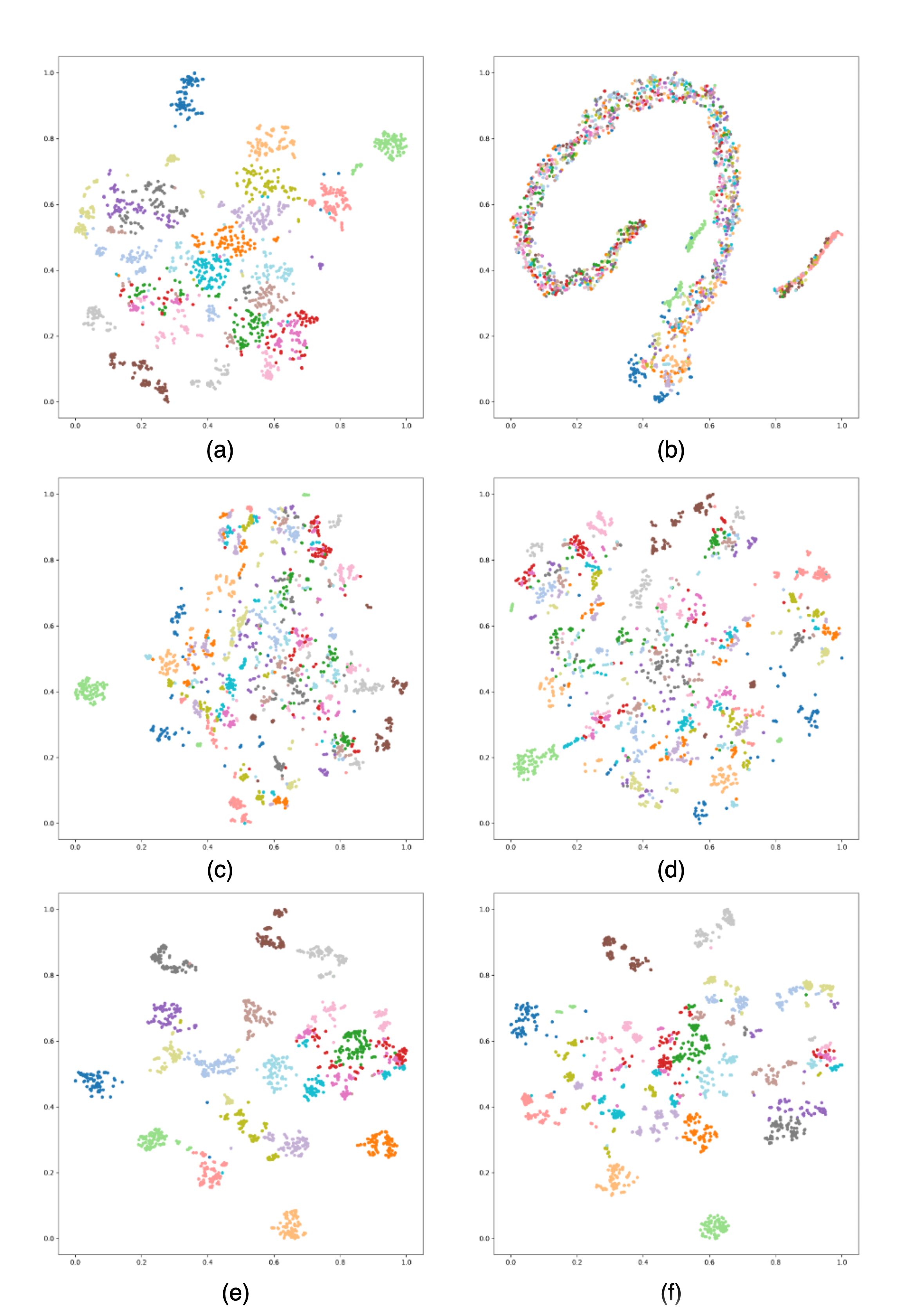}
    \vspace{-0.7cm}
    \caption{The t-SNE visualization results of representations extracted from different pre-trained models are shown in the figures (a) through (f). These results were obtained by extracting the output of the last layer of the pre-trained models (which were not fine-tuned) on the UCM dataset. The models shown are as follows: (a) Barlow-Twins, (b) SeCo, (c) ViT-B-RVSA, (d) ViTAE-B-RVSA, (e) CMID-ResNet50, and (f) CMID-Swin-B.}
    \label{fig:tsne}
\vspace{-0.5cm}
\end{figure}
\begin{figure}[t!]
    \centering
    \includegraphics[width=\linewidth]{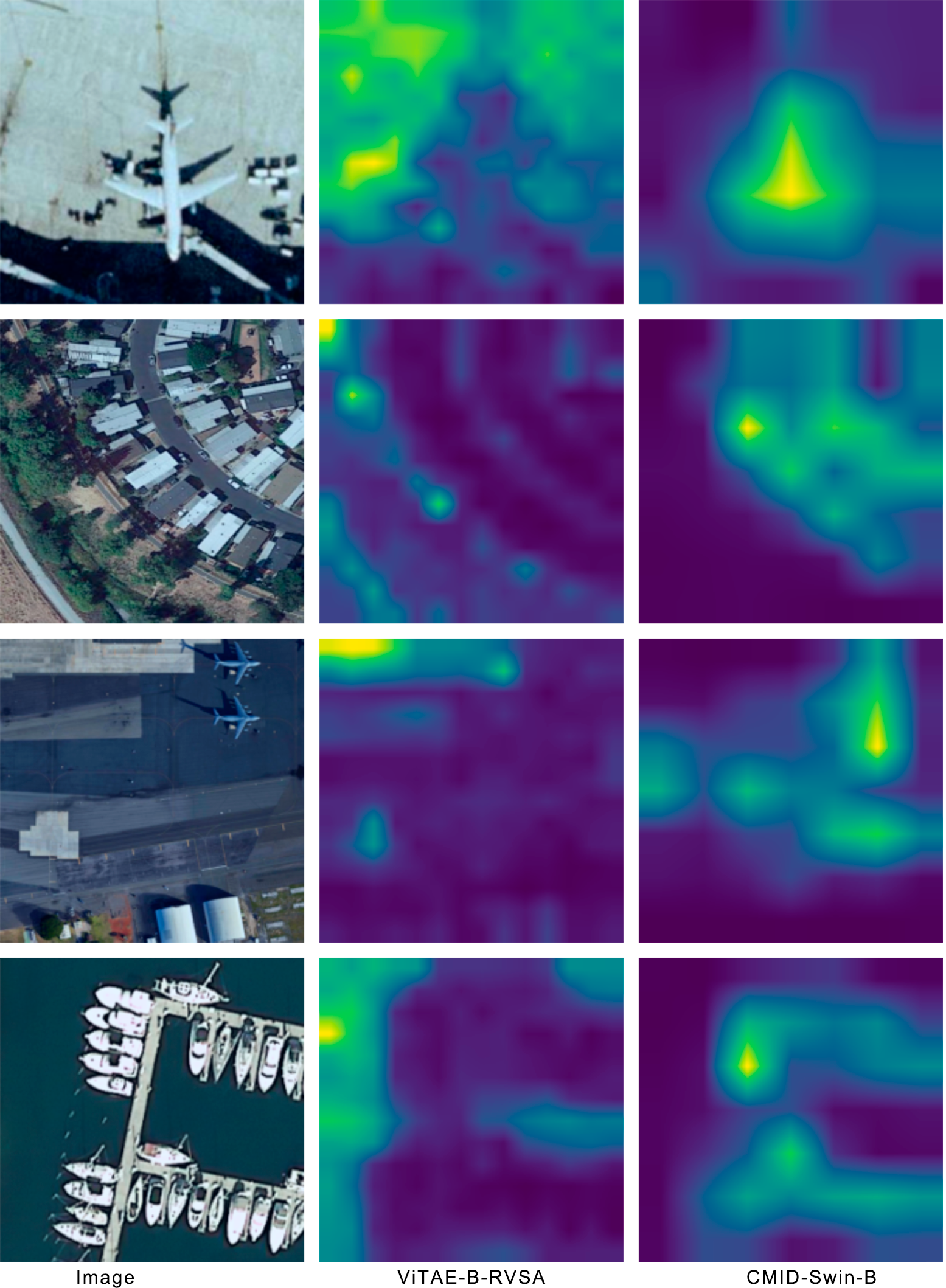}
    \vspace{-0.7cm}
    \caption{Visualization the self-attention map of the last layer of the pre-trained models by averaging the values of the heads of the last attention layer. These maps demonstrate that the CMID pre-trained model is able to effectively focus on the main objects in the image, rather than on other details, indicating its ability to learn and distinguish important representations. These models have not undergone any fine-tuning or been attached to any labels. }
    \label{fig:attentionmap}
\end{figure}

\subsubsection{Object Detection}
We visualize the failed OBB detection results on the DOTA dataset in Fig.~\ref{fig:failExample}~(b) following the same setting described in Sec.~\ref{sec:objectdetection}. 
The results indicate that ResNet-50 pre-trained by CMID performs inadequately in detecting small objects. This is likely due to the information loss issue when applying the MIM method to CNNs~\cite{jing2022masked,liu2022mixmim}. Although CMID effectively combines CL and MIM to allow the MIM method to be applied to CNNs, the inherent continuity of the convolution operation restricts it to regular grids and cannot handle irregular, non-overlapping patches such as visible and invisible ones, leading to a decrease in performance~\cite{tian2023designing}. The key to resolving this issue is to modify the convolution operation to handle irregular, non-overlapping patches, such as by implementing sparse convolution~\cite{liu2015sparse} or partial convolution~\cite{liu2018partial}.

\section{Discussion}
\subsection{The Characteristics Of The Representations Learned By CMID}
The primary goal of CMID is to combine MIM and CL to learn representations that are both global semantic separable and local spatial perceptible. The exceptional performance of CMID in various RS downstream tasks has shown that the representations learned by CMID indeed have these two characteristics. In this section, we will further investigate the representations learned by CMID using various visualization tools.

\begin{figure}[t!]
    \centering
    \includegraphics[width=1\linewidth]{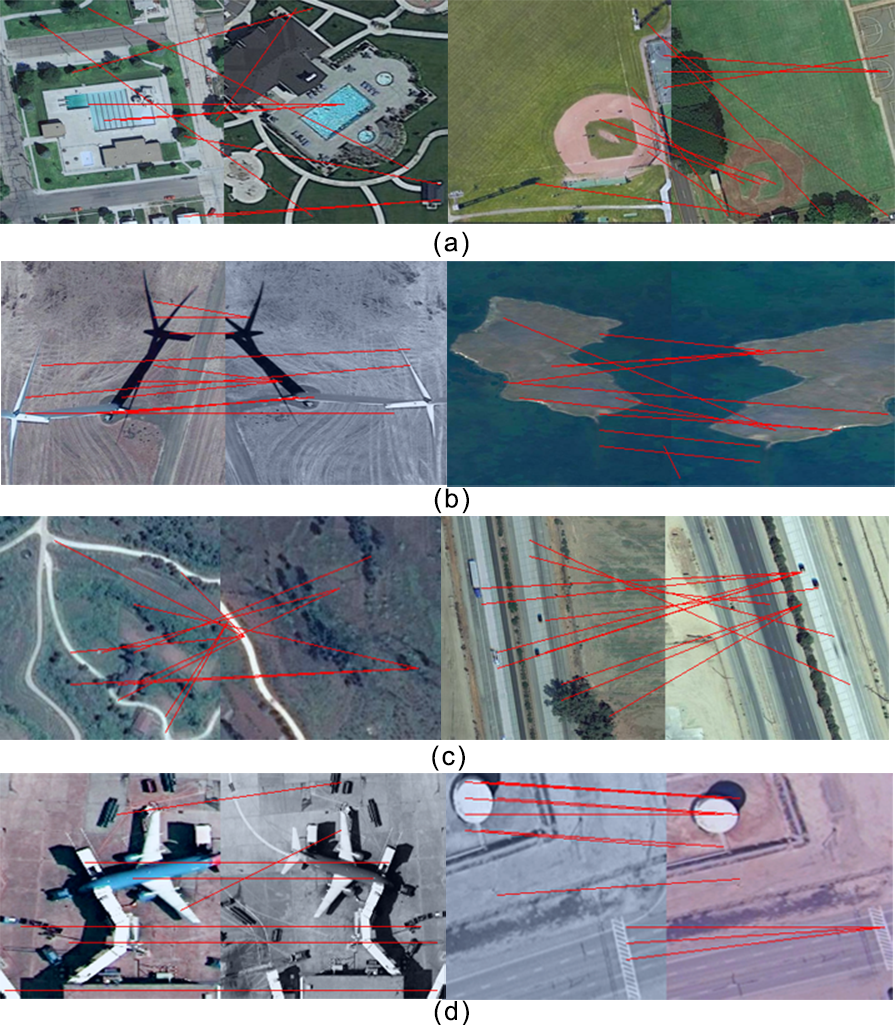}
    \vspace{-0.7cm}
    \caption{Visualization of feature correspondence. The correspondence is extracted using CMID pre-trained Swin-B (a, b) and ResNet-50 (c, d).  Image pairs in (a) and (c) are sampled from two images of the same class, while (b) and (d) are generated from two augmented views of a single image. These matching results demonstrate the strong ability of CMID to match features at a local scale despite variations in color and spatial locations.}
    \label{fig:correspond}
\vspace{-0.5cm}
\end{figure}
We start by visualizing the extracted representations on the UCM dataset using the t-SNE~\cite{hinton2002stochastic} method. As shown in Fig.~\ref{fig:tsne},  the extracted representations using the proposed CMID method (e and f) are better separated than those obtained from other SSL methods. It is worth noting that the representations extracted using only the MIM method (Fig.~\ref{fig:tsne} c and d) are not highly separable, but CMID is able to achieve more globally semantically separable representations by combining CL with MIM. It is also worth noting that the representations extracted by CMID-Swin-B are not as separable as those extracted by CMID-ResNet50. This may explain why CMID-Swin-B does not perform as well as CMID-ResNet50 with limited training data in the classification task (Fig.~\ref{clsresultfigure}). We expect that further adjustments to the CL learning strategy could enhance the separability of the representations extracted by ViTs.

In Fig.~\ref{fig:attentionmap}, we visualize the attention maps of the self-attention layers for models pre-trained with ViTAE-B-RVSA~\cite{wang2022advancing} and CMID-Swin-B. The results show that the ViTAE-B-RVSA pre-trained model focuses more on redundant and low-information regions, such as the apron around an aircraft (first row of Fig.~\ref{fig:attentionmap}) or the sea near a ship (fourth row of Fig.~\ref{fig:attentionmap}). This aligns with the MIM method's tendency to recover noise and low-level image details~\cite{wei2022masked,assran2022masked}. In contrast, the proposed CMID pre-trained model focuses more on semantically informative regions, which enables the model to more effectively locate the main object in images and improves its performance in downstream tasks such as segmentation and detection.

Finally, we visualize the most similar feature correspondences between two augmented views of the same image or two images labeled as the same class using the output feature map of the third stage of the pre-trained models (Fig.~\ref{fig:correspond}). 
The 12 correspondences with the highest similarity scores are shown. 
The results demonstrate that CMID is able to accurately match most correspondences even with large changes in color and scale (b and d). It also has a strong ability to distinguish between different objects without causing confusion, as seen in the precise matching of the same objects in different images of the same class (a and b). These results indicate the robustness of CMID's learned representations among different augmentations and its ability to discriminate between different objects.
\subsection{Expectation and Limitation}
Learning effective representations that are universal across various remote sensing visual understanding tasks, without human annotated labels in training, is a longstanding goal for analyzing RS images. SSL, with its ability to learn task-agnostic representations by effectively exploiting the huge quantity of unlabeled remote sensing data, has witnessed great success in the field of remote sensing~\cite{wang2022advancing,sun2022ringmo,manas2021seasonal,ayush2021geography}. Many studies have also explored to incorporating additional information to enrich the learned representations~\cite{yuan2022sits,cong2022satmae}. However, most existing remote sensing SSL methods are limited to learning either global semantic separable or local spatial perceptible representations, resulting in methods that may work well for a specific downstream task but not for others. 
We suppose that a successful SSL method should be able to improve the performance of the model for any downstream task and that should be unified in the sense that it can be applied to different network architectures and used with different-sized datasets. Along this direction, we propose CMID to unify the SSL framework for representation learning of RS images. 
We anticipate that the proposed CMID will be applied to a broader range of datasets and to more and newer network architectures in order to fully realize its potential.
We also expect that the unified learning framework will serve as a baseline for representation learning of RS images and lead to advances in the field of automatic interpretation of RS images.

The proposed CMID SSL has demonstrated impressive results, however, it still has some limitations. Firstly, CMID employs the SimMIM~\cite{xie2022simmim} method in the MIM branch to preserve the structure of the image, allowing the application of the MIM method to hierarchical architectures such as Swin-Transformer~\cite{liu2021swin} and ResNet~\cite{he2016deep}. However, the SimMIM encoder processes both visible and invisible patches, unlike the MAE~\cite{he2022masked} method, which only processes the visible patches, resulting in reduced learning efficiency compared to MAE. Thus, a more efficient MIM method that can be used for hierarchical architectures may be a viable alternative to the MIM branch of CMID. 
Secondly, while CMID successfully integrates CL and MIM to enable the application of MIM to CNNs, the inherent continuity of the convolution operation restricts it to regular grids, hindering its ability to handle irregular, non-overlapping patches, such as visible and invisible patches, leading to decreased performance~\cite{tian2023designing}. Thus, there is a need to modify the convolution operation to fully utilize the MIM learning signals and explore the potential of MIM pre-training on CNN. 
Finally, the representations extracted by CMID-Swin-B are less separable than those extracted by CMID-ResNet50 (Fig.~\ref{fig:tsne}). It is assumed that adapting the global branch of CMID to better suit the characteristics of ViT could result in improved global semantic separability of the learned representations.

\section{Conclusion}
We present CMID, a self-supervised learning framework for representation learning of RS images. CMID combines the advantages of contrastive learning and masked image modeling to learn representations with both global semantic separability and local spatial perceptibility in a self-distillation manner. Through extensive experiments on various downstream remote sensing tasks, we demonstrate that CMID significantly enriches the learned representations and outperforms state-of-the-art self-supervised methods on multiple downstream tasks. Additionally, our CMID framework is architecture-agnostic, allowing it to be easily applied to both convolutional neural networks and vision transformers for a wide range of DL applications in RS images interpretation. We expect CMID will serve as a baseline for representation learning of RS images and lead to advances in the field of automatic interpretation of RS images.

%

\section*{Acknowledgment}
We are grateful to High Performance Computing Center of Nanjing University for their help on GPU resources.
We also would like to thank the editor and the anonymous reviewers for there constructive comments.

\ifCLASSOPTIONcaptionsoff
  \newpage
\fi



\bibliographystyle{IEEEtran}
\normalem
\bibliography{bibtex/bib/ref}

%








\end{document}